\documentclass[accepted]{uai2025} 
                        
\usepackage[american]{babel}
\usepackage{comment}

\usepackage{natbib} 
\usepackage{mathtools} 
\usepackage{booktabs} 

\usepackage{wrapfig}
\usepackage{graphicx}
\usepackage{amsmath}
\usepackage{amssymb}
\usepackage{amsthm}
\usepackage{float}
\usepackage{subfig}
\usepackage[table]{xcolor}

\usepackage[ruled,linesnumbered]{algorithm2e}

\usepackage[all=normal, mathspacing=normal,mathdisplays=normal, floats=tight, paragraphs=tight, lists=normal]{savetrees}

\newcommand{\X}{\mathcal{X}}
\newcommand{\Y}{\mathcal{Y}}
\newcommand{\D}{\mathcal{D}}
\newcommand{\R}{\mathbb{R}}
\newcommand{\Z}{\mathcal{Z}}

\newcommand{\rvx}{\mathbf{x}}
\newcommand{\rvy}{\mathbf{y}}
\newcommand{\rvz}{\mathbf{z}}
\newcommand{\rve}{\mathbf{e}}
\newcommand{\vx}{\boldsymbol{x}}
\newcommand{\vy}{\boldsymbol{y}}
\newcommand{\vz}{\boldsymbol{z}}
\newcommand{\veps}{\boldsymbol{\epsilon}}
\newcommand{\ve}{\boldsymbol{e}}

\newcommand{\vg}{\boldsymbol{g}}

\definecolor{mplblue}{rgb}{0.12156862745098039, 0.4666666666666667, 0.7058823529411765}
\definecolor{mplorange}{rgb}{1.0, 0.4980392156862745, 0.054901960784313725}
\definecolor{mplred}{rgb}{0.8392156862745098, 0.15294117647058825, 0.1568627450980392}
\definecolor{mplgreen}{rgb}{0.17254901960784313, 0.6274509803921569, 0.17254901960784313}
\definecolor{mplgrey}{rgb}{0.4980392156862745, 0.4980392156862745, 0.4980392156862745}
\usepackage{tikz}
\usetikzlibrary{bayesnet, arrows, tikzmark, calc, matrix, fit, patterns}
\usepackage{scalerel}
\newsavebox{\boxblue}
\newsavebox{\boxorange}
\newsavebox{\boxgrey}
\sbox{\boxblue}{\tikz \draw[mplblue!100, line width=6pt] (-0.4,-0.4) -- (0.4,0.4) (-0.4,0.4) -- (0.4,-0.4);}
\sbox{\boxorange}{\tikz \fill[mplorange!100] (0,0) circle (1.0);}

\title{Generative Uncertainty in Diffusion Models}

\author[1]{Metod Jazbec\thanks{Corresponding author: $<$m.jazbec@uva.nl$>$}}
\author[2]{Eliot Wong-Toi}
\author[3]{Guoxuan Xia}
\author[4]{Dan Zhang}
\author[5]{Eric Nalisnick}
\author[2]{Stephan Mandt}

\affil[1]{%
    UvA-Bosch Delta Lab,
    University of Amsterdam \quad $^2$University of California, Irvine 
} 
\affil[3]{Imperial College, London \quad $^4$Bosch Center for AI \quad $^5$Johns Hopkins University}

\begin{document}
\maketitle

\begin{abstract}
  Diffusion models have recently driven significant breakthroughs in generative modeling. While state-of-the-art models produce high-quality samples \textit{on average}, individual samples can still be low quality. Detecting such samples without human inspection remains a challenging task. To address this, we propose a Bayesian framework for estimating \textit{generative uncertainty} of synthetic samples.  We outline how to make Bayesian inference practical for large, modern generative models and introduce a new semantic likelihood (evaluated in the latent space of a feature extractor) to address the challenges posed by high-dimensional sample spaces. Through our experiments, we demonstrate that the proposed generative uncertainty effectively identifies poor-quality samples and significantly outperforms existing uncertainty-based methods.  Notably, our Bayesian framework can be applied \textit{post-hoc} to any pretrained diffusion or flow matching model (via the Laplace approximation), and we propose simple yet effective techniques to minimize its computational overhead during sampling. 
\end{abstract}

\section{Introduction}\label{sec:intro}

Diffusion (and flow-matching) models \citep{sohl2015deep, song2020denoising, song2020score, lipman2022flow} have recently pushed the boundaries of generative modeling due to their strong theoretical underpinnings and scalability. Across various domains, they have enabled the generation of increasingly realistic samples \citep{Rombach_2022_CVPR, esser2024scaling,li2024absorb}. Despite the impressive progress, state-of-the-art models can still generate low quality images that contain artefacts and fail to align with the provided conditioning information. This poses a challenge for deploying diffusion models, as it can lead to a poor user experience by requiring multiple generations to manually find an artefact-free sample.

Bayesian inference has long been applied to detect poor-quality predictions in predictive models \citep{mackay1992information, gal2016uncertainty, wilson2020case, arbel2023primer}. By capturing the uncertainty of the model parameters due to limited training data, each prediction can be assigned a \textit{predictive uncertainty}, which, when high, serves as a warning that the prediction may be unreliable. Despite its widespread use for principled uncertainty quantification in predictive models, Bayesian methodology has been far less commonly applied to detecting poor generations in generative modeling. 
This raises a key question: \textit{How can Bayesian principles help us detect poor generations?}

In this work, we propose a Bayesian framework for estimating \textit{generative uncertainty} in modern generative models, such as diffusion. To scale Bayesian inference for large diffusion models, we employ the (last-layer) Laplace approximation \citep{mackay1992bayesian, ritter2018scalable, daxberger2021laplace}. Additionally, to address the challenge posed by the high-dimensional sample spaces of data such as natural images, we introduce a semantic likelihood, where we leverage pretrained image encoders (such as CLIP \citep{radford2021learning}) to compute variability in a latent, \textit{semantic} space instead. Through our experiments, we demonstrate that generative uncertainty is an effective tool for detecting low-quality samples and propose simple strategies to minimize the sampling overhead introduced by Bayesian inference. In particular, we make the following contributions:

\begin{enumerate}
    \item We formalize the notion of \textit{generative uncertainty} and propose a method to estimate it for modern generative models (Section~\ref{sec:methods}). Analogous to how predictive uncertainty helps identify unreliable predictions in predictive models, generative uncertainty can be used to detect low-quality generations in generative models.
    \item We show that our generative uncertainty strongly outperforms previous uncertainty-based approaches for filtering out poor samples \citep{kou2023bayesdiff, de2024diffusion} (Section~\ref{sec:exp-poor}). Additionally, we achieve competitive performance with non-uncertainty-based methods, such as realism score  \citep{kynkaanniemi2019improved} and rarity score \citep{han2022rarity}, while also highlighting the complementary benefits of uncertainty (Appendix \ref{app:corr}).
    \item We propose effective strategies to reduce the sampling overhead of Bayesian uncertainty (Section~\ref{sec:exp-eff}) and demonstrate the applicability of our framework beyond diffusion models by applying it to a (latent) flow matching model (Section~\ref{app:flow-match}).
\end{enumerate}





\section{Background}

\subsection{Generative Modeling}
\label{back:gen-model}
\paragraph{Sampling in Generative Models} Modern deep generative models like variational autoencoders (VAEs) \citep{kingma2013auto}, generative adversarial networks (GANs) \citep{goodfellow2014generative}, and diffusion models differ in their exact probabilistic frameworks and training schemes, yet share a common sampling recipe: start with random noise and transform it into a new data sample \citep{tomzcak2022book}. Specifically, let $\vx \in \X$ denote a data sample and $\vz \in \Z$ an initial noise. A new sample is generated by:
\begin{align*}
\vz \sim p(\rvz)\:, \;\;\; \hat{\vx} = g_{\theta}(\vz) \: ,
\end{align*}
where $p(\rvz)$ is an initial noise (prior) distribution, typically a standard Gaussian $\mathcal{N}(0, I)$, and $g_{\theta}: \Z \rightarrow \X$ is a generator function with model parameters $\theta \in \R^P$. Here and throughout the paper, we use $\vz$ (with a slight abuse of notation) to denote the entire randomness involved in the sampling process.\footnote{This distinction matters for diffusion models. In DDIM \citep{song2020denoising} and ODE sampling \citep{song2020score}, randomness is only present at the start of the sampling process (akin to VAEs and GANs). In contrast, in DDPM \citep{ho2020denoising} and SDE sampling \citep{song2020score}, randomness is introduced at every step throughout the sampling process.}


\paragraph{Diffusion Models} The primary focus of this work is on diffusion models \citep{sohl2015deep}. These models operate by progressively corrupting data into Gaussian noise and learning to reverse this process. For a data sample $\vx_0 \sim q(\rvx)$, the forward (noising) process is defined as
\begin{align*}
\vx_t = \sqrt{\bar{\alpha}_t} \vx_0 + \sqrt{1 - \bar{\alpha_t}} \veps, \; \veps \sim \mathcal{N}(0, I)
\end{align*}
where $\bar{\alpha}_t = \prod_{s=1}^t (1 - \beta_s)$ and $\{\beta_s\}_{s=1}^T$, with $\beta_S\in (0, 1)$, is a noise schedule chosen such that $\vx_T \sim \mathcal{N}(0, I)$ (approximately). In the backward process, a denoising network $f_{\theta}$ is learned via a simplified regression objective (among various  possible parameterizations, see \cite{song2020score} or \cite{karras2022elucidating}):
\begin{align}
\label{eq:diff-loss}
 \!  \! \!  \mathcal{L}(\theta) = \mathbb{E}_{t, \vx_0, \veps}  \! \left[ \big| \big |f_{\theta}(\sqrt{\bar{\alpha}_t} \vx_0 \!  +  \! \sqrt{1 \!  -  \! \bar{\alpha_t}} \veps, t) - \veps   \big| \big |_2^2 \right] \: .
\end{align}
 After training, diffusion models generate new samples via a generator function, $g_{\hat{\theta}}$, which consists of sequentially applying the learned denoiser, $f_{\hat{\theta}}$, and following specific transition rules from samplers such as DDPM \citep{ho2020denoising} or DDIM \citep{song2020denoising}.

\begin{figure*}[t]
\vspace{-5mm}
    \centering\includegraphics[width=0.85\textwidth]{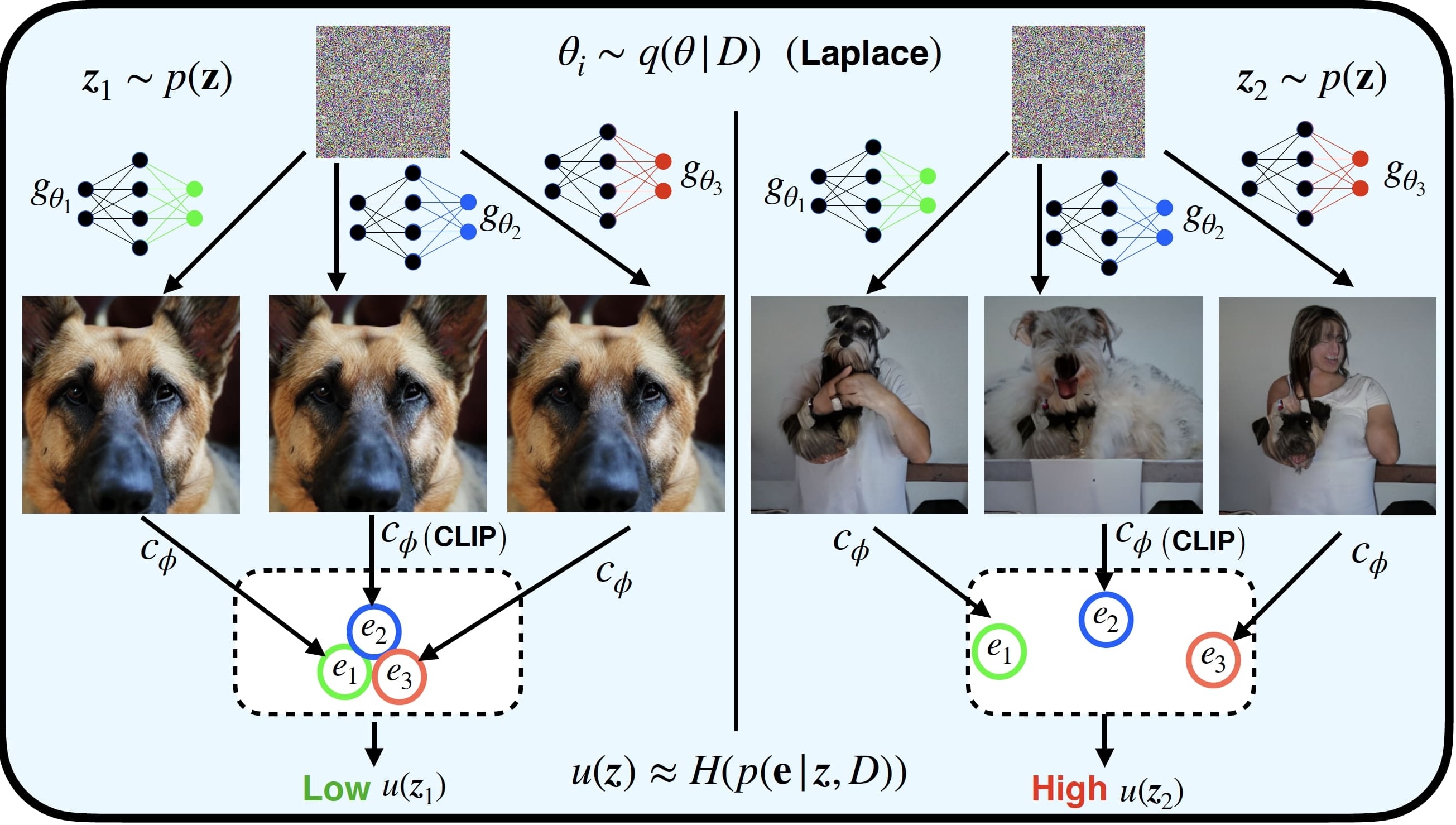}

    \caption{
    \textbf{Generative uncertainty.}
    Illustration of how we compute generative uncertainty for a fixed random noise $\vz$. For a diffusion model $g_{\theta}$, we draw $M$ parameter sets $\{\theta_1,…, \theta_M\}$ from the approximate posterior $q(\theta | D)$ over diffusion model's parameters (here: last-layer Laplace with $M=3$). Each model $g_{\theta_m}$ maps $\vz$ to an image $\hat{\vx}_m$. Embedding all $\hat{\vx}_m$ with a frozen encoder $c_\phi$ (e.g., CLIP) gives $M$ semantic feature vectors $\ve_m$; the variability (e.g., entropy) of these vectors is the final uncertainty $u(\vz)$. Low-uncertainty (left) corresponds to consistent, high-quality images, whereas high-uncertainty (right) reveals model disagreement and poor, discordant outputs. 
    }
    \label{fig:methods}
\end{figure*}

\subsection{Bayesian Deep Learning}
\label{back:bayes}
Bayesian neural networks (BNNs) go beyond point predictions and allow for principled uncertainty quantification \citep{neal2012bayesian, kendall2017uncertainties, jospin2022hands}. Let $h_{\psi}: \mathcal{X} \rightarrow \mathcal{Y}$ denote a predictive model with parameters $\psi \in \R^O$ and $\mathcal{D} = \{(\vx_n, \vy_n) \}_{n=1}^N$ denote training data. Instead of finding a single fixed set of parameters, $\hat{\psi} = \arg \max \mathcal{L}(\psi ; \mathcal{D})$, that maximizes a chosen objective function $\mathcal{L}$, BNNs specify a prior $p(\psi)$ over model parameters and define a likelihood $p(\rvy | h_{\psi}(\vx))$, which together yield a posterior distribution via Bayes rule: $p(\psi | \mathcal{D}) \propto p(\psi) \prod_{n=1}^N p(\vy_n | h_{\psi}(\vx_n))$. 
Under this Bayesian view, a predictive model for a new test point $\vx_*$ is then obtained via the posterior predictive distribution:
\begin{align*}
p(\rvy | \vx_*, \mathcal{D}) = \mathbb{E}_{p(\psi | \mathcal{D})} \big[p(\rvy | h_{\psi}(\vx_*))\big]. \: 
\end{align*}
For large models, finding the exact posterior distribution is computationally intractable, hence an approximate posterior $q(\psi | \mathcal{D})$ is used instead. Popular approaches for approximate inference include deep ensembles \citep{lakshminarayanan2017simple, wilson2020bayesian}, variational inference \citep{blundell2015weight, zhang2018advances}, SWAG \citep{mandt2017stochastic, maddox2019simple}, and Laplace approximation \citep{daxberger2021laplace}. Moreover, to alleviate computational overhead, it is common to give a `Bayesian treatment' only to a subset of parameters \citep{kristiadi2020being, daxberger2021bayesian, sharma2023bayesian}. Finally, the intractable expectation integral in the posterior predictive is approximated via Monte-Carlo (MC) sampling:
\begin{align}
\label{eq:bma-approx}
p(\rvy | \vx_*, \mathcal{D}) \!  \approx    \! \frac{1}{M} \!  \! \sum_{m=1}^M p(\rvy | h_{\psi_m} \! (\vx_*)), \: \; \psi_m  \! \sim  \! q(\psi | \mathcal{D}), \:
\end{align}
with $M$ denoting the number of MC samples. By measuring the variability of the posterior predictive distribution, e.g., its entropy, one can obtain an estimate of the model's predictive uncertainty for a given test point $u(\vx_*)$. The utility of such uncertainties has been demonstrated on a wide range of tasks such as out-of-distribution (OOD) detection \citep{daxberger2021laplace}, active learning \citep{gal2017deep}, and detection of influential samples \citep{nickl2024memory}.

\section{Generative Uncertainty via Bayesian Inference}
\label{sec:methods}

While Bayesian neural networks (BNNs) have traditionally been applied to predictive models to estimate \textit{predictive uncertainty}, in this section we demonstrate how to apply them to diffusion to estimate \textit{generative uncertainty}  (see Figure~\ref{fig:methods} and Algorithm \ref{algo:diff-gen-unc} for an overview of our method). Later in Section~\ref{sec:exp}, we show that generative uncertainty can be used to detect poor-quality samples. Our focus is on generative models for natural images, where $\vx \in \R^{H \times W \times C}$. For ease of exposition, we consider unconditional generation in this section, though our methodology can also be applied directly to conditional models (see Section~\ref{sec:exp-poor}).

\subsection{Generative Uncertainty} As in traditional Bayesian predictive models (cf. Section~\ref{back:bayes}), the central principle for obtaining a Bayesian notion of uncertainty in diffusion models is the posterior predictive distribution:
\begin{align}
\label{eq:gen-BMA}
p(\rvx | \vz, \mathcal{D}) = \mathbb{E}_{p(\theta | \mathcal{D})} \big[p(\rvx | g_{\theta}(\vz))\big]. \:
\end{align}
Here, as before in Section \ref{back:gen-model} , we use $\vz$ to denote the entire randomness involved in the diffusion sampling process. Generative uncertainty is then defined as the variability of the posterior predictive:
\begin{align}
\label{eq:gen-unc}
u(\vz) := \mathcal{V}(p(\rvx | \vz, \mathcal{D}))
\end{align}
where $\mathcal{V}(\cdot)$ denotes the variability measure, such as entropy. We propose a tractable estimator of the posterior predictive later in Eq.~\ref{eq:gen-bma-approx}. 

In the same way that the predictive uncertainty $u(\vx_*)$, of a predictive model provides insight into the quality of its prediction for a new test point $\vx_*$, the generative uncertainty $u(\vz)$ of a generative model $g_{\theta}$ should offer information about the quality of the generation $g_{\theta}(\vz)$ for a `new' random noise sample $\vz$. We demonstrate this relationship experimentally in Section~\ref{sec:exp}. Next, we discuss how to make Bayesian inference on (large) diffusion models computationally tractable.

\subsection{Last-Layer Laplace Approximation}
\label{sec:methods-llla}
State-of-the-art diffusion models are extremely large (100M to 1B+ parameters) and can take weeks to train. Consequently, the computational overhead of performing Bayesian inference on such large models is of significant concern. For instance, while deep ensembles are a convenient and popular approach for predictive models \citep{lakshminarayanan2017simple}, the sheer size of diffusion models renders naive ensembling infeasible. To address this, we adopt the Laplace approximation \citep{mackay1992bayesian, shun1995laplace} to find the approximate posterior $q(\theta | \D)$. The Laplace approximation is among the most computationally efficient approximate inference methods while still offering competitive performance \citep{daxberger2021laplace}. Moreover, a particularly appealing feature of the Laplace approximation is that it can be applied \textit{post-hoc} to any diffusion model. We leverage this property in Section~\ref{sec:exp}, where we apply it to a variety of popular diffusion and flow-matching models.

The Laplace approximation of the posterior is given by:
\begin{align}
\label{eq:la}
q(\theta | \D) = \mathcal{N}(\theta | \hat{\theta}, \Sigma), \; \Sigma = \big(\nabla_{\theta}^2 \mathcal{L}(\theta ; \D) \big|_{\hat{\theta}} \big)^{-1}  ,
\end{align}
where $\hat{\theta}$ represents the parameters of a pre-trained diffusion model, and $\Sigma$ is the inverse Hessian of the diffusion training loss from Eq.~\ref{eq:diff-loss}. To reduce the computational cost further, we apply a `Bayesian' treatment only to the last layer of the denoising network $f_{\theta}$. 

Note that for the Laplace approximation to be theoretically valid, the loss function should correspond to (log-)likelihood and (log-)prior terms. While the diffusion loss can be interpreted as a log-likelihood up to an additive constant—due to its role as a surrogate for the KL divergence between the noising and denoising processes—this interpretation holds only under appropriate weighting of the loss terms across different time steps \citep{song2021maximum}. Consequently, applying the Laplace approximation directly, without such reweighting, is not fully theoretically justified. Despite this, we find in our experiments (Section~\ref{sec:exp-poor}) that applying the Laplace approximation directly to the diffusion losses used in practice (without modifying the loss weighting) still yields meaningful uncertainty estimates that align well with the visual quality of generated samples. That said, we believe that better understanding the theoretical requirements for applying approximate Bayesian inference techniques like Laplace in modern generative models like diffusion remains an important direction for future work that could lead to even more informative uncertainty estimates.

It is also worth noting that the use of last-layer Laplace approximation for diffusion models has been previously proposed in BayesDiff \citep{kou2023bayesdiff}. While our implementation of the Laplace approximation closely follows theirs, there are significant differences in how we utilize the approximate posterior, $q(\theta | \D)$. Specifically, in our approach, we use it within the traditional Bayesian framework (Eq.~\ref{eq:gen-BMA}) to sample new diffusion model parameters, leaving the diffusion sampling process, $g_{\theta}$, unchanged. In contrast, BayesDiff resamples new weights from $q(\theta | \D)$ at every diffusion sampling step $t$, which necessitates substantial modifications to the diffusion sampling process through their \textit{variance propagation} approach. We later demonstrate in Section~\ref{sec:exp-poor} that modifications such as variance propagation are unnecessary for obtaining Bayesian generative uncertainty and staying closer to the traditional Bayesian setting leads to the best empirical performance.

\begin{algorithm}[t]
\caption{Diffusion Sampling  with Generative Unc.\label{algo:diff-gen-unc}}
\DontPrintSemicolon
\SetAlgoLined
\LinesNumbered
\SetKwInOut{Input}{Input}
\SetKwInOut{Output}{Output}

\Input{random noise $\vz$, pretrained diffusion model $g_{\hat{\theta}}$, Laplace posterior $q(\theta|\mathcal{D})$ (Eq.~\ref{eq:la}), number of MC samples $M$, semantic feature extractor $c_{\phi}$, semantic likelihood noise $\sigma$}
\Output{generated sample $\hat{\vx}_0$, generative uncertainty estimate $u(\vz)$}

Generate a sample $\hat{\vx}_0 = g_{\hat{\theta}}(\vz)$\;
Get semantic features $\ve_0 = c_{\phi}(\hat{\vx}_0)$\;

\For{$m = 1 \rightarrow M$}{
    $\theta_m \sim q(\theta|\mathcal{D})$\;
    $\hat{\vx}_m = g_{\theta_m}(\vz)$\;
    $\ve_m = c_{\phi}(\hat{\vx}_m)$\;
}

Compute $p(\vx|\vz,\mathcal{D})$ using $\{\ve_m\}_{m=0}^M$ (Eq.~\ref{eq:gen-bma-approx})\;
Compute the entropy $u(\vz) = H(p(\vx|\vz,\mathcal{D}))$\;

\Return{$\hat{\vx}_0, u(\vz)$}
\end{algorithm}

\subsection{Semantic Likelihood}
\label{sec:sem-lik}
We next discuss the choice of likelihood for estimating generative uncertainty in diffusion models. Since the denoising problem in diffusion is modeled as a (multi-output) regression problem, the most straightforward approach is to place a simple Gaussian distribution over the generated sample:
\begin{align}
\label{eq:pixel-likelihood}
p(\rvx | g_{\theta}(\vz)) = \mathcal{N}(\rvx \: |\:  g_{\theta}(\vz), \sigma^2 I),
\end{align}
where $\sigma^2$ represents the observation noise.

\begin{figure*}[t]
\vspace{-5mm}
    \centering\includegraphics[width=0.99\textwidth]{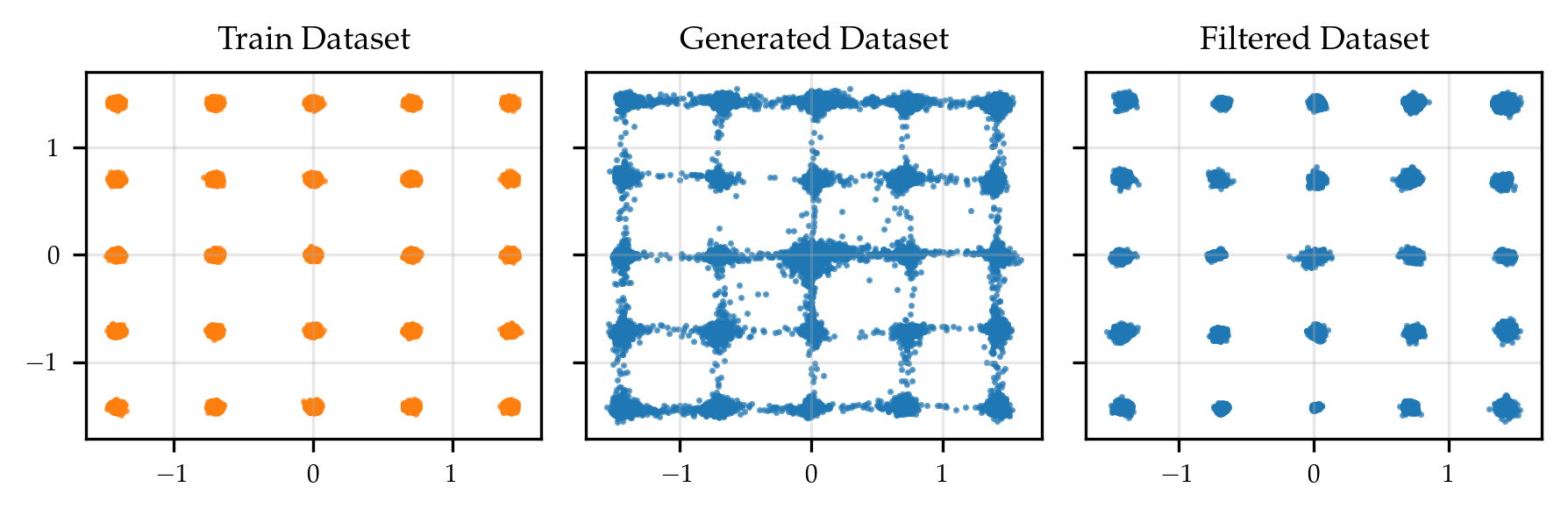}
    \caption{Illustration of how generative uncertainty can be effective for filtering our poor generations on a simple 2D Gaussian dataset. \emph{Left:} training data consisting of 25 separate Gaussian modes. \emph{Middle:} generated samples using a trained diffusion model and a DDPM sampler ($T=1000$). \emph{Right:} same set of generated samples after removing 50\% of generations with the highest estimated generative uncertainty.}
    \label{fig:toy-data}
\end{figure*}

However, as we will demonstrate in Section~\ref{sec:exp}, this likelihood leads to non-informative estimates of generative uncertainty (Eq.~\ref{eq:gen-unc}). The primary issue is that the sample space of natural images is high-dimensional (i.e., $|\X| = HWC$). Consequently, placing the likelihood directly in the sample space causes the variability of the posterior predictive distribution to be based on pixel-level differences. This is problematic because it is well-known that two images can appear nearly identical to the human eye while exhibiting a large $L_2$-norm difference in pixel space $\X$ (see, for example, the literature on adversarial examples \citep{szegedy2013intriguing}). To get around this, we propose to map the generated samples to a `semantic' latent space, $\mathcal{S}$, via a pre-trained feature extractor, $c_{\phi}: \X \rightarrow \mathcal{S}$ (e.g., an inception-net \citep{szegedy2016rethinking} or a CLIP encoder \citep{radford2021learning}). The resulting \textit{semantic likelihood} has the form
\begin{align}
\label{eq:sem-lik}
    p(\rvx | g_{\theta}(\vz) ; \phi) = \mathcal{N}(\rve(\rvx) \: | \: c_{\phi}\big(g_{\theta}(\vz)\big), \sigma^2 I)
\end{align}
where $\rve(\rvx) \in \mathcal{S}$ is the (random) vector of semantic features. By combining the (last-layer) Laplace approximate posterior and the semantic likelihood, we can now approximate the posterior predictive (Eq.~\ref{eq:gen-BMA}) as
\begin{align}
\label{eq:gen-bma-approx}
    &p(\rvx | \vz, \mathcal{D}) \approx \mathcal{N}\big(\rve(\rvx) \: \big| \: \bar{\ve}, \: \textrm{Diag}\big(\frac{1}{M}\sum_{m=1}^M \ve_m ^ 2 - \bar{\ve}^2 \big) + \sigma^2 I  \big) \: , \nonumber \\ 
     &\bar{\ve} = \frac{1}{M} \sum_{m=1}^M \ve_m, \;\;\; \ve_m \! = c_{\phi}\big(g_{\theta_m}(\vz)\big), \: \theta_m  \! \sim q(\theta | \D),  
\end{align}
where $M$ denotes the number of Monte Carlo samples. Additionally, we approximate the posterior predictive with a single Gaussian via moment matching here, a common practice in Bayesian neural networks for regression problems \citep{lakshminarayanan2017simple, antoran2020depth}. A more detailed derivation is provided in Appendix~\ref{app:gen-unc-approx}.

Unlike in the posterior predictive for predictive models (Eq.~\ref{eq:bma-approx}), where it is used to obtain both the prediction and the associated uncertainty, the generative posterior predictive (Eq.~\ref{eq:gen-bma-approx}) is used solely to estimate the generative uncertainty $u(\vz)$. The actual  samples $\hat{\vx}$ are still generated using the pre-trained diffusion model $g_{\hat{\theta}}$ (see Algorithm~\ref{algo:diff-gen-unc}). As a variability measure $\mathcal{V}(\cdot)$ in our generative uncertainty framework, we propose to use entropy (denoted with $H(\cdot)$ in Algorithm~\ref{algo:diff-gen-unc}) due to its simplicity and widespread use in quantifying predictive uncertainty. However, we note that alternative measures of variability, such as pairwise-distance estimators (PAiDEs) \citep{berry2023escaping}, can also be employed.


\section{Experiments}
\label{sec:exp}

In our experiments, we begin by demonstrating that generative uncertainty serves as an effective method for identifying poor samples in diffusion models, using a simple synthetic dataset (Section~\ref{sec:exp-toy}). We then show that our proposed approximations (Sections~\ref{sec:methods-llla} and~\ref{sec:sem-lik}) enable the estimation of generative uncertainty in large, modern diffusion models applied to high-dimensional natural images (Section~\ref{sec:exp-poor}). Additionally, we discuss the sampling overhead introduced by our Bayesian approach and show that it can be effectively reduced (Section~\ref{sec:exp-eff}). Finally, we extend our Bayesian framework beyond diffusion by applying it to detect low-quality samples in a (latent) flow matching model (Appendix~\ref{app:flow-match}). Our code is available at \url{https://github.com/metodj/DIFF-UQ}. 

\subsection{Toy Demonstration}
\label{sec:exp-toy}
To illustrate the potential of generative uncertainty for detecting poor generations, we adopt the setting from \citet{aithal2024understanding}. Specifically, we use a 2D synthetic dataset with 25 distinct modes (Figure~\ref{fig:toy-data}, \emph{left}) to train a small diffusion model. We then generate 50K samples using a DDPM sampler (Figure~\ref{fig:toy-data}, \emph{middle}). While the generated samples cover the 25 modes well, many `hallucinated' samples also appear between the modes of the training data. Following \citet{aithal2024understanding}, we consider such samples to be poor generations, as they are highly unlikely under the true data-generating distribution.

Next, we train an ensemble of diffusion models ($M=5$) and use it to estimate the generative uncertainty of each of the 50K generated samples. We then filter out the 50\% of samples with the highest estimated uncertainty and plot the remaining ones in Figure~\ref{fig:toy-data}, \emph{right}. As shown in the plot, this uncertainty-based filtering effectively removes all poor generations between modes, indicating that generative uncertainty can serve as a reliable indicator of sample quality. Note that in this simple toy setting, we use neither the Laplace approximation (relying instead on a diffusion deep ensemble) nor the semantic likelihood. In the following section, we show how both can be employed to extend generative uncertainty estimation to the more realistic setting of natural images.

\subsection{Detecting Low-Quality Generations}
\label{sec:exp-poor}

\begin{figure*}[htbp]
\vspace{-5mm}
    \centering\includegraphics[width=0.99\textwidth]{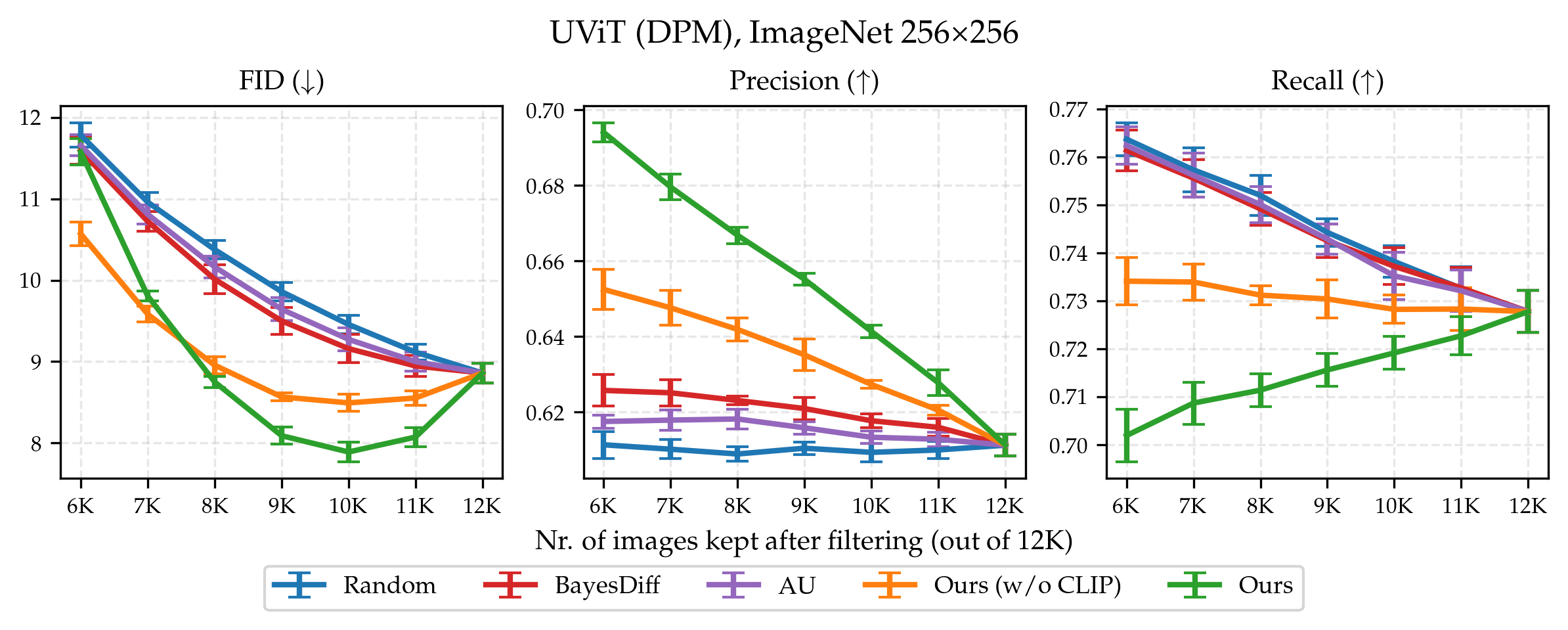}

    \caption{Image generation results for $n \in \{6\textrm{K}, 7\textrm{K}, \ldots, 11\textrm{K} \} $ filtered samples (out of 12K) for UViT diffusion model \citep{bao2023all}. Our generative uncertainty outperforms previously proposed uncertainty-based approaches (AU \citep{de2024diffusion}, BayesDiff \citep{kou2023bayesdiff}) in terms of image quality, as indicated by higher FID (\emph{left}) and precision (\emph{middle}) scores. We report mean values along with standard deviations over 5 runs with different random seeds.}
    \label{fig:unc-filter-uvit}
\end{figure*}

To demonstrate that our proposed generative uncertainty is effective for detecting low-quality generations also on high-dimensional data such as natural images, we follow the experimental setup from prior work on uncertainty-based filtering \citep{kou2023bayesdiff, de2024diffusion}. Specifically, we generate 12K samples using a given diffusion model and compute the uncertainty estimate for each sample. We then select $n \in \{6\textrm{K}, 7\textrm{K}, \ldots, 11\textrm{K} \}$  samples with the \textit{lowest} uncertainty. If uncertainty reliably reflects the visual quality of generated samples, filtering based on it should yield greater improvements in population-level metrics (such as FID) compared to selecting a random subset of $n$ images.

\paragraph{Implementation Details} To ensure a fair comparison with BayesDiff \citep{kou2023bayesdiff}, we adopt their proposed implementation of the last-layer Laplace approximation. Specifically, we use an Empirical Fisher approximation of the Hessian with a diagonal factorization \citep{daxberger2021laplace}. When computing the posterior predictive distribution (Eq.~\ref{eq:gen-bma-approx}), we use $M = 5$ Monte Carlo samples. For the semantic feature extractor $c_{\phi}$, we leverage a pretrained CLIP encoder \citep{radford2021learning}. We set the observation noise to $\sigma^2=0.001$ (Eq. \ref{eq:sem-lik}). Additional implementation details are provided in Appendix~\ref{app:impl}.

\textbf{Baselines} We compare our proposed generative uncertainty to existing uncertainty-based approaches for detecting low-quality samples: BayesDiff and the aleatoric uncertainty (AU) approach proposed by \cite{de2024diffusion}. BayesDiff estimates epistemic uncertainty in diffusion models using a last-layer Laplace approximation and tracks this uncertainty throughout the entire sampling process. In contrast, in AU, uncertainty is computed by measuring the sensitivity of intermediate diffusion scores to random perturbations. Unlike our approach, both methods estimate uncertainty directly in pixel space.

\paragraph{Evaluation Metrics} In addition to the widely used Fréchet Inception Distance (FID) \citep{heusel2017gans} for evaluating the quality of a filtered set of images, we also report \textit{precision} and \textit{recall} metrics \citep{sajjadi2018assessing, kynkaanniemi2019improved}. To compute these quantities we fit two manifolds in feature space: one for the generated images and another for the reference (training) images. Precision is the proportion of generated images that lie in the reference image manifold, while recall is the proportion of reference images that lie in the generated image manifold. Precision measures the quality (or fidelity) of generated samples, whereas recall quantifies their diversity (or coverage over the reference distribution).

\paragraph{Results} We present our main results on the ImageNet dataset in Figure~\ref{fig:unc-filter-uvit} for UViT model \citep{bao2023all} and in Figure~\ref{fig:unc-filter-adm} for ADM model \citep{dhariwal2021diffusion}. We first observe that existing uncertainty-based approaches (BayesDiff and AU) result in little to no improvement in metrics that assess sample quality (FID and precision). In contrast, our generative uncertainty method leads to significant improvements in terms of both FID and precision. For example, on the UViT model, a subset of $n=10\textrm{K}$ images selected based on our uncertainty measure achieves an FID of $7.89$, significantly outperforming both the Random baseline ($9.45$) and existing uncertainty-based methods (BayesDiff $9.16$, AU $9.20$).

\begin{figure*}[htbp]
    \centering
    \begin{minipage}{0.47\textwidth}
        \centering
        \includegraphics[width=\textwidth]{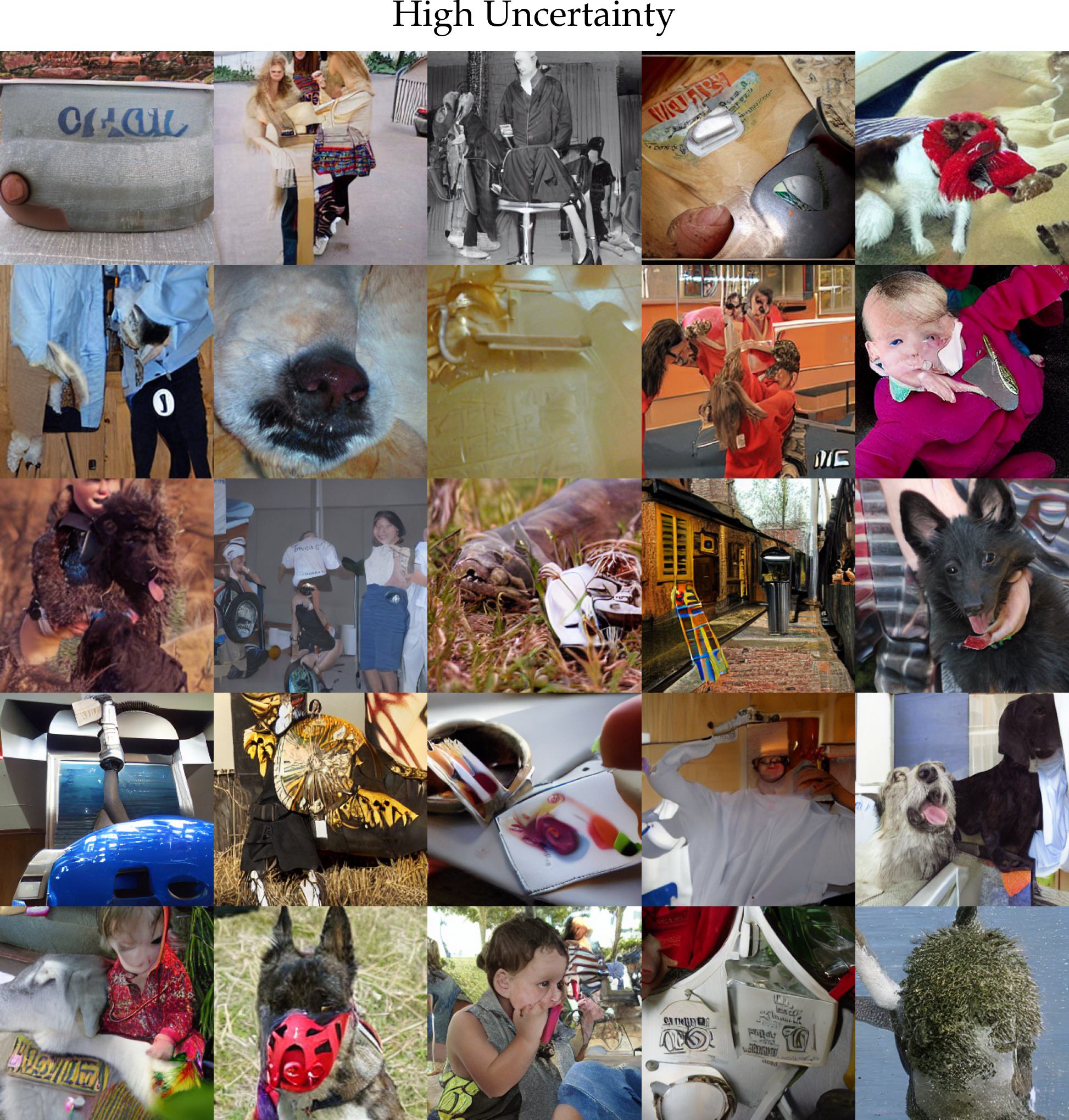}
    \end{minipage}
    \hfill
    \begin{minipage}{0.47\textwidth}
        \centering
        \includegraphics[width=\textwidth]{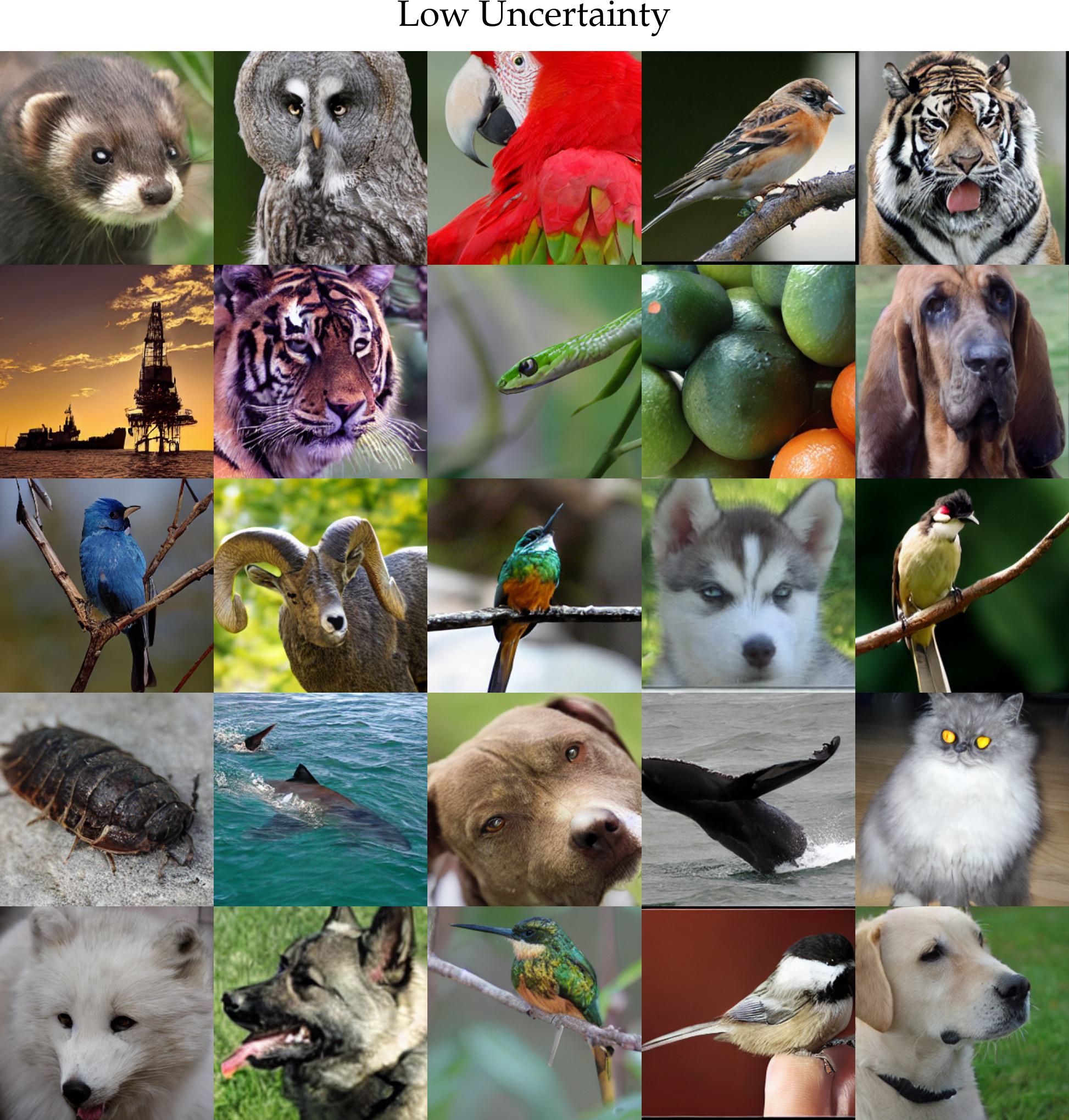}
    \end{minipage}
    \caption{Images with highest (\textit{left}) and lowest  (\textit{right}) generative uncertainty amongst 12K generations using a UViT diffusion model \citep{bao2023all}. Generative uncertainty correlates with visual quality: high-uncertainty samples exhibit numerous artefacts, whereas low-uncertainty samples resemble canonical images of their respective conditioning class.}
    \label{fig:best-worst-ours}
\end{figure*}

Next, in order to qualitatively demonstrate the effectiveness of our approach, we show 25 samples with the highest and lowest generative uncertainty (out of the original 12K samples) according to our method in Figure~\ref{fig:best-worst-ours}. High-uncertainty samples exhibit numerous artefacts, and in most cases, it is difficult to determine what exactly they depict. Combined with the quantitative results in Figures~\ref{fig:unc-filter-uvit}\&\ref{fig:unc-filter-adm}, this supports our hypothesis that (Bayesian) generative uncertainty is an effective metric for identifying low-quality samples. Conversely, the lowest-uncertainty samples are of high quality, with most appearing as `canonical' examples of their respective (conditioning) class.

For comparison, in Figure~\ref{fig:best-worst-bayesdiff} we also depict the 25 `worst' and `best' samples according to the uncertainty estimate from BayesDiff. It is evident that their uncertainty is less informative for sample quality than ours. Moreover, their uncertainty measure appears to be very sensitive to the background pixels. Most images with the highest uncertainty have a `cluttered' background, whereas most images with the lowest uncertainty have a `clear' background. We attribute this issue to the fact that in BayesDiff the uncertainty is computed directly in the pixel space, unlike in our approach where we use the semantic likelihood (Section~\ref{sec:sem-lik}) to move away from the (high-dimensional) sample space. To further verify the importance of the semantic likelihood,  we perform an ablation where we compute the generative uncertainty directly in the pixel-space (see {\tikz[baseline=-0.5ex] \draw[mplorange!100, very thick, solid] (0., 0.) -- (0.3, 0.);} lines). As seen in Figures~\ref{fig:unc-filter-uvit}\&\ref{fig:unc-filter-adm}, this leads to worse FID/precision numbers in most cases compared to using the proposed semantic likelihood. Moreover, based on samples in Figure~\ref{fig:best-worst-ours-no-cphi}, it is clear that without semantic likelihood, our uncertainty becomes overly sensitive to the background pixels in the same way as in BayesDiff.

\begin{figure*}[htbp]
    \centering
    \begin{minipage}{0.47\textwidth}
        \centering
        \includegraphics[width=\textwidth]{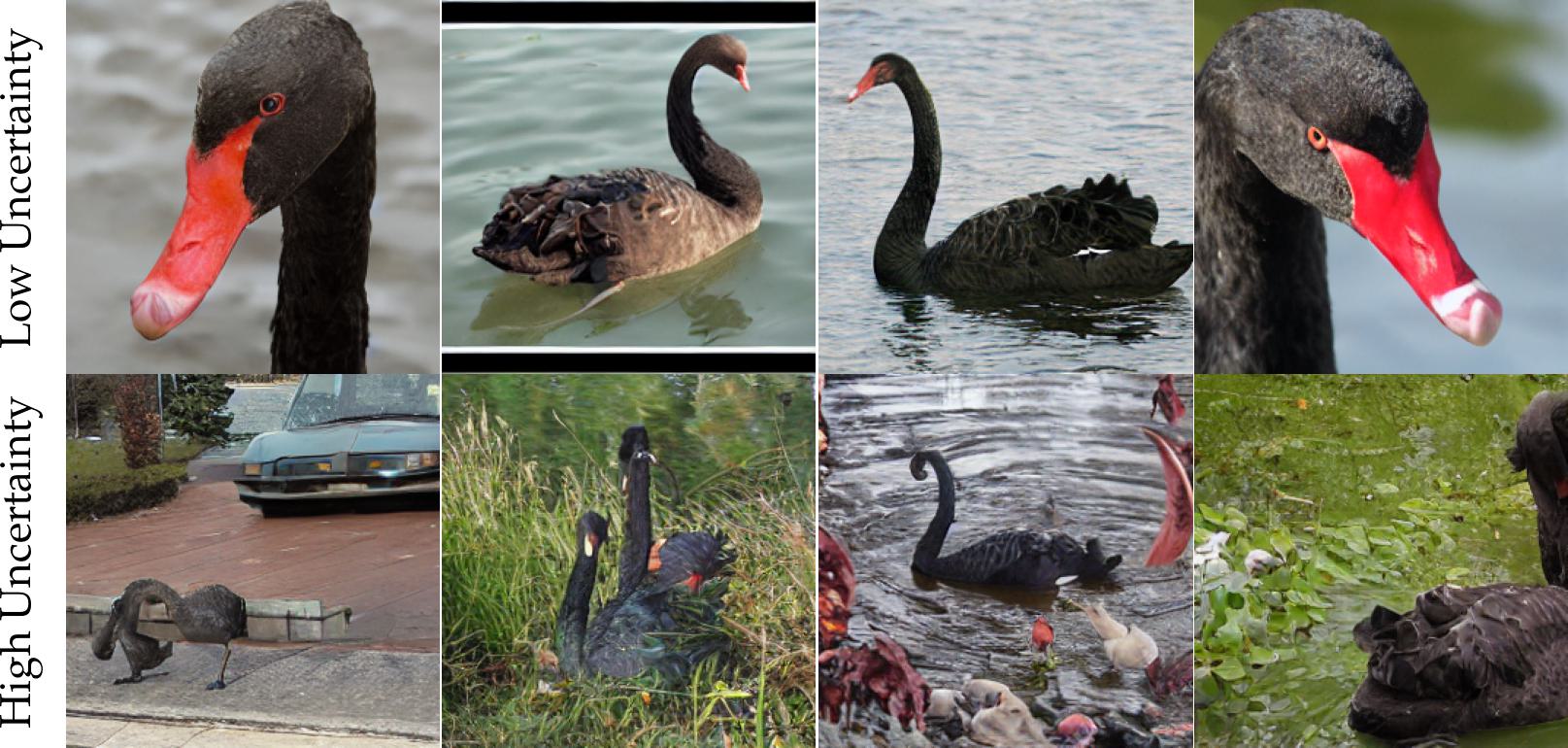}
    \end{minipage}
    \hfill
    \begin{minipage}{0.47\textwidth}
        \centering
        \includegraphics[width=\textwidth]{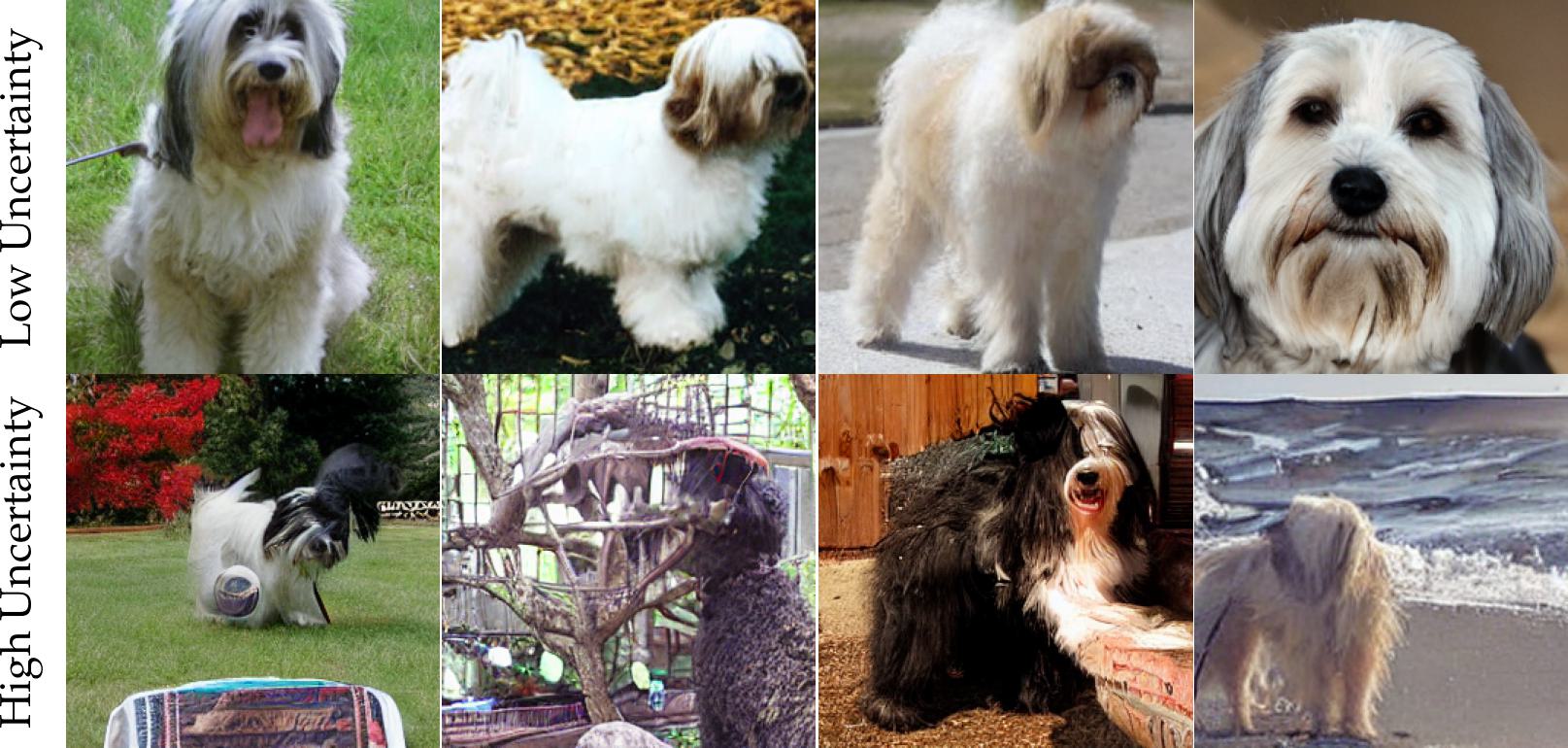}
    \end{minipage}
    \caption{Images with the highest (\textit{bottom}) and the lowest (\textit{top}) generative uncertainty among 128 generations using a UViT diffusion model for 2 classes: \texttt{black swan} (\textit{left}) and \texttt{Tibetan terrier} (\textit{right}).}
    \label{fig:best-worst-ours-cond}
\end{figure*}

Lastly, we observe that filtering based on our generative uncertainty results in some loss of sample diversity, as evidenced by lower recall scores (see \emph{right} plots in Figures~\ref{fig:unc-filter-uvit}\&\ref{fig:unc-filter-adm}). We attribute this to the fact that, in our main experiment, 12K images are generated such that all 1000 ImageNet classes are represented.\footnote{Following \cite{kou2023bayesdiff}, we use class-conditional diffusion models but randomly sample a class for each of the 12K generated samples.} Since certain classes produce images with higher uncertainty (see Appendix~\ref{app:class-ent} for a detailed analysis), filtering based on uncertainty inevitably alters the class distribution among the selected samples. 
Moreover, the trade-off between improving sample quality (precision) and reducing diversity (recall) has been observed before, see for example the literature on classifier(-free) guidance \citep{ho2022classifier}.

\paragraph{Comparison with realism and rarity scores} Lastly, we compare our proposed method with non-uncertainty-based approaches, such as the \textit{realism} score \citep{kynkaanniemi2019improved} and the \textit{rarity} score \citep{han2022rarity}. These metrics work by measuring the distance of a generated sample from the data manifold (derived from a reference dataset) in a semantic space spanned by the inception-net features \citep{szegedy2016rethinking}. Notably, prior work \citep{kou2023bayesdiff, de2024diffusion} has not considered such comparisons, which we believe are essential for assessing the practical utility of uncertainty-based filtering.

For realism, we retain the $n$ images with the highest scores, whereas for rarity, we keep those with the lowest scores. As shown in Figures~\ref{fig:unc-filter-adm-realism-rarity}\&\ref{fig:unc-filter-uvit-realism-rarity}, our generative uncertainty performs on par with realism and rarity in terms of FID. However, compared to our method, realism and rarity scores result in sharper precision-recall trade-offs—yielding larger precision gains at the expense of a greater drop in recall.

Furthermore, Table~\ref{tab:combined-scores} shows that our score can be effectively combined with realism or rarity scores. Specifically for $n=10\textrm{K}$, combining our score with realism yields an FID of $7.60$ on UViT, compared to $8.26$ when combining realism and rarity. We attribute higher benefits from ensembling our score to the fact that, while realism and rarity exhibit a strong negative Spearman correlation (-0.85), our uncertainty measure is less correlated with them (-0.27 with realism, 0.38 with rarity), as shown in Figure~\ref{fig:metrics-corr}. Taken together, these results indicate that our uncertainty score captures (somewhat) different desirable properties of images compared to realism and rarity.


\subsection{Improving Sampling Efficiency}
\label{sec:exp-eff}

We next examine the sampling costs associated with Bayesian inference in diffusion sampling. As shown in Algorithm~\ref{algo:diff-gen-unc}, obtaining an uncertainty estimate $u(\vz)$ for a generated sample $\hat{\vx}_0 = g_\theta(\vz)$ requires generating $M$ additional samples, resulting in 
$MT$ additional network function evaluations (NFEs). For the results presented in Section~\ref{sec:exp-poor}, we use $M=5$ and the default number of sampling steps $T=50$ (\protect\scalerel*{\usebox{\boxblue}}{\circ}), leading to an additional 250 NFEs for uncertainty estimation—on top of the 50 NFEs required to generate the original sample. Since this overhead may be prohibitively expensive in certain deployment scenarios, we next explore strategies to reduce the sampling cost associated with our generative uncertainty.

\begin{figure}[t]
    \centering
    \includegraphics[width=0.95\linewidth]{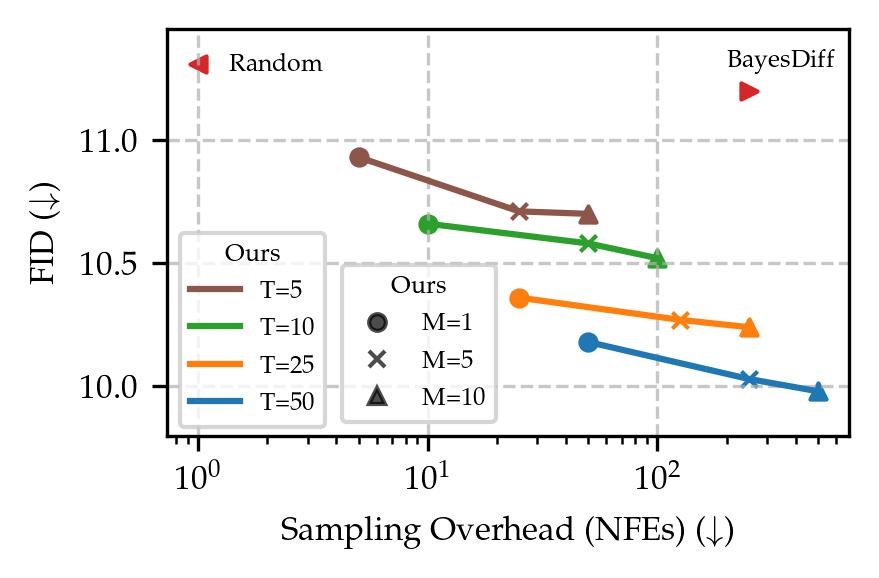}
    \caption{FID results for $n=10\textrm{K}$ ImageNet filtered images using our generative uncertainty on ADM model \citep{dhariwal2021diffusion}. We vary the number of Monte Carlo samples $M$ and diffusion sampling steps $T$ (see Algorithm~\ref{algo:diff-gen-unc}). By default, we use $M$=5 with $T$=50 (\protect\scalerel*{\usebox{\boxblue}}{\square}), incurring an additional 250 NFEs for uncertainty estimation. Encouragingly, setting $M$=1 and $T$=25 (\protect\scalerel*{\usebox{\boxorange}}{\square}) still achieves competitive performance while reducing the sampling overhead by $10$x. Lower left is best: better FID and greater computational efficiency.}
    \label{fig:efficiency-fid}
    \vspace{-1\baselineskip}
\end{figure}

The most straightforward approach is to reduce the number of Monte Carlo samples $M$. Encouragingly, reducing $M$ to as few as 1 still achieves highly competitive performance (see Figure~\ref{fig:efficiency-fid}). Further efficiency gains can be achieved by reducing the number of sampling steps $T$, leveraging the flexibility of diffusion models to adjust $T$ on the fly. Importantly, we lower $T$ only for the additional $M$ samples used for uncertainty assessment while keeping the default $T$ for the original sample $\hat{\vx}_0$ to ensure that the generation quality is not compromised. Taken together, reducing $M$ and $T$ significantly improves the efficiency of our generative uncertainty. Using the ADM model \citep{dhariwal2021diffusion}, our generative uncertainty method with $M=1$ and $T=25$ (\protect\scalerel*{\usebox{\boxorange}}{\circ}) achieves an FID of 10.36, which still strongly outperforms both the Random ($11.31$) and BayesDiff ($11.20$) baselines while requiring only $25$ additional NFEs.


\section{Related Work}






\textbf{Uncertainty quantification in diffusion} models has recently gained significant attention. Most related to our work are BayesDiff \citep{kou2023bayesdiff}, which uses a Laplace approximation to track epistemic uncertainty throughout the sampling process, and \cite{de2024diffusion}, which captures aleatoric uncertainty via the sensitivity of diffusion score estimates. Our work extends both by introducing an uncertainty framework that is more general (applicable beyond diffusion), simpler (requiring no sampling modifications), and more effective (see Section~\ref{sec:exp-poor}).

Also related is DECU \citep{berryshedding}, which employs an efficient variant of deep ensembles \citep{lakshminarayanan2017simple} to capture the epistemic uncertainty of conditional diffusion models. However, DECU does not consider using uncertainty to detect poor-quality generations, as its framework provides uncertainty estimates at the level of the conditioning variable, whereas ours estimates uncertainty at the level of initial random noise. Similarly, in \cite{chanestimating} the use of hyper-ensembles is proposed to capture epistemic uncertainty in diffusion models for inverse problems such as super-resolution, but, as in DECU, their approach does not provide uncertainty estimates in unconditional settings or in conditional settings with low-dimensional conditioning (such as class-conditional generation). Moreover, both DECU \citep{berryshedding} and \cite{chanestimating} require modifying and retraining diffusion model components, whereas our approach operates \textit{post-hoc} with any pretrained diffusion model via the Laplace approximation \citep{daxberger2021laplace}. A recent approach, PUNC \citep{franchi2024unct2i}, focuses only on text-to-image models. The uncertainty of image generation with respect to text conditioning is measured through the alignment between a caption generated from a generated image and the original prompt used to generate said image.

Additionally, a large body of work explores conformal prediction for uncertainty quantification in diffusion models \citep{ angelopoulos2022image, sankaranarayanan2022semantic, teneggi2023trust, belhasin2023principal}. However, these approaches are primarily designed for inverse problems (e.g., deblurring), and cannot be directly applied to detect low-quality samples in unconditional generation.


\textbf{Bayesian inference in generative models} has been explored previously outside the domain of diffusion models. Prominent examples include \cite{saatci2017bayesian} where a Bayesian version of a GAN is proposed, showing improvements for semi-supervised learning, and \cite{daxberger2019bayesian}, where a Bayesian VAE~\citep{tran2023fully} is shown to provide more informative likelihood estimates for the unsupervised out-of-distribution detection compared to the non-Bayesian counterparts \citep{nalisnick2018deep}. Since diffusion models can be interpreted as neural ODEs \citep{song2020score}, another relevant work is \cite{ott2023uncertainty}, which employs a Laplace approximation to quantify uncertainty when solving neural ODEs \citep{chen2018neural}. However, \cite{ott2023uncertainty} focuses solely on low-dimensional regression problems. 

\textbf{Non-uncertainty based approaches for filtering out poor generations} include the realism \citep{kynkaanniemi2019improved}, rarity \citep{han2022rarity}, and anomaly scores \citep{hwang2024anomaly}. Our work is the first to establish a connection between these scores and uncertainty-based methods, which we hope will inspire the development of even better sample-level metrics in the future. Additionally, a large body of work focuses on specially designed sample-quality scoring models \citep{gu2020giqa, zhao2024lightweight} or, alternatively, on leveraging large pretrained vision-language models (VLMs) \citep{zhang2024bench} for scoring generated images. However, these approaches require either access to sample-quality labels or rely on (expensive) external VLMs. In contrast, our uncertainty-based method requires neither, making it a more accessible and scalable alternative.

\section{Limitations}
\label{sec:limit}

While we have demonstrated in Section~\ref{sec:exp} that semantic likelihood is essential for addressing the over-sensitivity of prior work to background pixels \citep{kou2023bayesdiff}, our reliance on a pretrained image encoder like CLIP \citep{radford2021learning} limits the applicability of our diffusion uncertainty framework to natural images. Removing the dependence on such encoders would unlock the application our Bayesian framework to other modalities where diffusion models are used, such as molecules \citep{hoogeboom2022equivariant, cornet2024equivariant} or text \citep{gong2022diffuseq, yi2024diffusion}. Exploring whether insights from the literature on uncovering semantic features in diffusion models \citep{kwon2022diffusion, luo2024diffusion, namekata2024emerdiff} could help achieve this represents a promising direction for future work.

Moreover, the large size of modern diffusion models necessitates the use of cheap and scalable Bayesian approximate inference techniques, such as the (diagonal) last-layer Laplace approximation employed in our work (following \citep{kou2023bayesdiff}). A more comprehensive comparison of available approximate inference methods could be valuable, as improving the quality of the posterior approximation may further enhance the detection of low-quality samples based on Bayesian generative uncertainty.


\section{Conclusion} We introduced generative uncertainty and demonstrated how to estimate it in modern generative models such as diffusion. Our experiments showed the effectiveness of generative uncertainty in filtering out low-quality samples. For future work, it would be interesting to explore broader applications of Bayesian principles in generative modeling beyond detecting poor-quality generations. Promising directions include guiding synthetic data generation and detecting memorized samples. It would also be worthwhile to further investigate the connection between uncertainty-based filtering and classifier(-free) guidance \citep{ho2022classifier}, as both exhibit similar precision-recall trade-offs.




\begin{acknowledgements} 
    We thank our reviewers for their thoughtful feedback. This project was generously supported by the Bosch Center for Artificial Intelligence. E.N. did not utilize resources from Johns Hopkins University for this project. S.M. acknowledges funding from the National Science Foundation (NSF) through an NSF CAREER Award IIS-2047418, IIS-2007719, the NSF LEAP Center, the IARPA WRIVA program, and the Hasso Plattner Research Center at UCI. See Appendix for additional disclaimers. 

\end{acknowledgements}


\bibliography{uai2025}

\newpage

\onecolumn

\section*{Appendix}

\appendix

The supplementary material is organized as follows:
\begin{itemize}
    \item In Appendix~\ref{app:figs}, we provide additional figures.
    \item In Appendix~\ref{app:gen-unc-approx}, we provide more details on our approximation of generative uncertainty (Eq. \ref{eq:gen-bma-approx}).
    \item In Appendix~\ref{app:exp-bayesdiff}, we qualitatively compare our method with BayesDiff \citep{kou2023bayesdiff}.
    \item In Appendix~\ref{app:exp-semlik}, we perform qualitative ablations on our semantic likelihood (Section~\ref{sec:sem-lik}).
    \item In Appendix~\ref{app:pixel}, we demonstrate how to use our generative uncertainty for pixel-wise uncertainty.
    \item In Appendix~\ref{app:llhood}, we show that diffusion's own likelihood is not useful for filtering out poor samples.
    \item In Appendix~\ref{app:corr}, we compare our generative uncertainty to realism \citep{kynkaanniemi2019improved} and rarity \citep{han2022rarity} scores.
    \item In Appendix~\ref{app:class-ent}, we investigate the drop in sample diversity by looking at the average generative uncertainty per conditioning class. 
    \item In Appendix~\ref{app:flow-match}, we  apply our generative uncertainty to detect low-quality samples in a latent flow matching model \citep{dao2023flow}.
    \item In Appendix~\ref{app:impl}, we provide implementation and experimental details. 
\end{itemize}

\section{Additional Figures}
\label{app:figs}
\begin{figure*}[htbp]
\vspace{-5mm}
    \centering\includegraphics[width=0.99\textwidth]{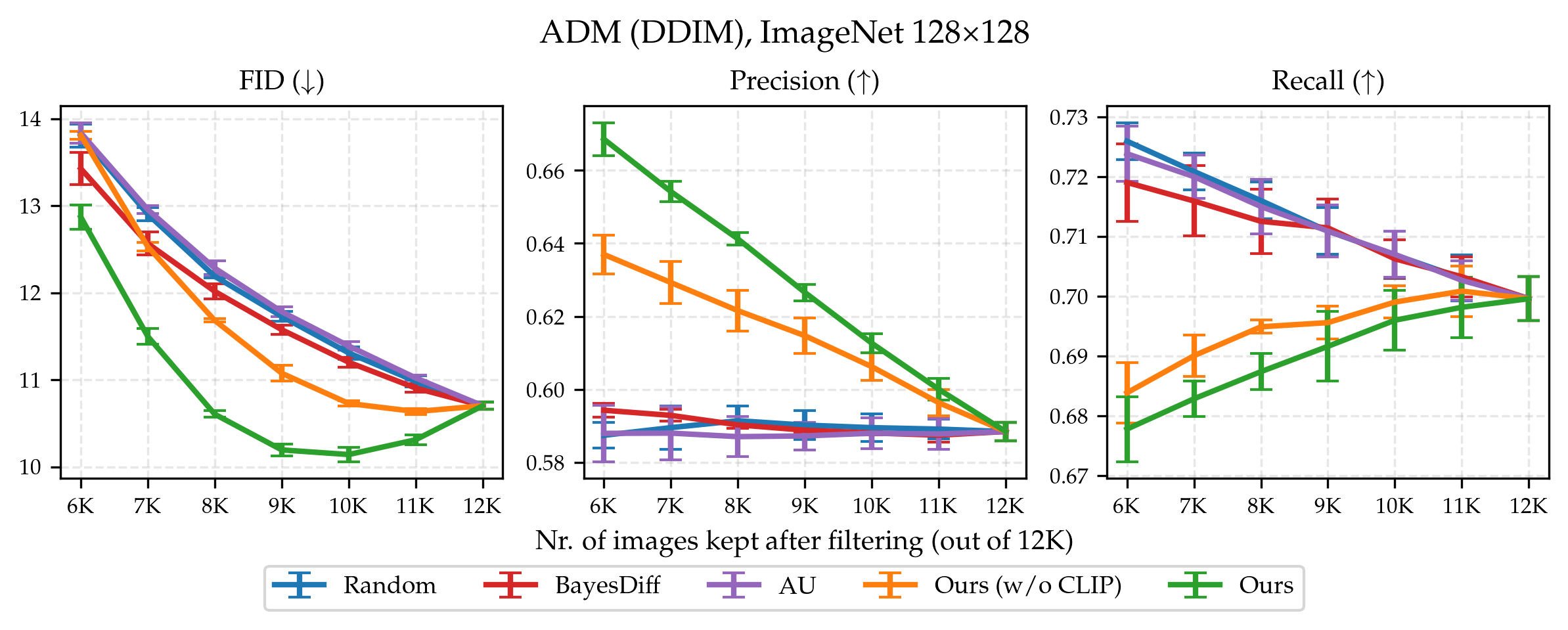}

    \caption{Image generation results for $n \in \{6\textrm{K}, 7\textrm{K}, \ldots, 11\textrm{K} \} $ filtered samples (out of 12K) for ADM diffusion model \citep{dhariwal2021diffusion}. Our generative uncertainty outperforms previously proposed uncertainty-based approaches (AU \citep{de2024diffusion}, BayesDiff \citep{kou2023bayesdiff}) in terms of image quality, as indicated by higher FID (\emph{left}) and precision (\emph{middle}) scores. We report mean values along with standard deviations over 5 runs with different random seeds.}
    \label{fig:unc-filter-adm}
\end{figure*}

\newpage
\section{Derivations}
\subsection{Generative Uncertainty Approximation}
\label{app:gen-unc-approx}
To ensure maximal clarity of exposition, we first derive a tractable estimator of  $u(\vz)$ (Eq. \ref{eq:gen-unc}) based on ‘pixel-space’ likelihood (Eq. \ref{eq:pixel-likelihood}):
\begin{align*}
p(\rvx|\vz, \mathcal{D}) \overset{Eq. \ref{eq:gen-BMA}}{=} \mathbb{E}_{p(\theta | \mathcal{D})} \big[p(\rvx | g_{\theta}(\vz)\big] \overset{(1)}{\approx} \frac{1}{M}\sum_{m=1}^M p(\rvx | g_{\theta_m}(\vz))  \overset{Eq. \ref{eq:pixel-likelihood}}{=}  \frac{1}{M}\sum_{m=1}^M \mathcal{N}(\rvx | g_{\theta_m}(\vz), \sigma^2 I) \overset{(2)}{\approx} \\ \mathcal{N}\big(\rvx| \bar{\vg}, \textrm{Diag}\big(\frac{1}{M}\sum_{m=1}^M g_{\theta_m}(\vz)^2 - \bar{\vg}^2\big) + \sigma^2 I\big) =: q_{\textrm{pixel}}(\vz)
\end{align*}
where $\theta_m \sim q(\theta | \mathcal{D})$ (Eq. \ref{eq:la}) and $\bar{\vg} = \frac{1}{M}\sum_{m=1}^M g_{\theta_m}(\vz)$. Note that in step $(1)$ we make use of the usual Monte Carlo (MC) approximation of the Bayesian posterior predictive (Eq. \ref{eq:bma-approx}) using $M$ samples and in step $(2)$ we approximate a mixture of Gaussian with a single Gaussian using moment-matching. Moreover, we consider only the diagonal of the resulting covariance which we do for efficiency reasons (e.g., for ImageNet 256x256 the full covariance has $\sim 4 \cdot 10^{10}$ parameters rendering the diagonal approximation necessary). 

A pixel-space generative uncertainty is then obtained by taking the entropy of the resulting Gaussian distribution: $u(\vz) \approx H(q_{\textrm{pixel}}(\vz))$. However, as we show qualitatively in Appendix~\ref{app:exp-semlik}, uncertainty based on pixel-space likelihood is not particularly informative about the visual quality of the samples as it is overly sensitive to the background pixels—images with simple backgrounds exhibit low uncertainty, whereas images with ‘cluttered’ background exhibit high uncertainty. This motivates our use of ‘semantic-likelihood’ (Eq. \ref{eq:sem-lik}) to arrive at the Eq. \ref{eq:gen-bma-approx}:
\begin{align*}
    p(\rvx | \vz, \mathcal{D})  
\overset{(1)}{\approx} \frac{1}{M}\sum_{m=1}^M p(\rvx | g_{\theta_m}(\vz) ; \phi ) 
\overset{\text{Eq.~\ref{eq:sem-lik}}}{=} \frac{1}{M}\sum_{m=1}^M \mathcal{N}(\ve(\rvx) | c_{\phi}(g_{\theta_m}(\vz)), \sigma^2 I) 
\overset{(2)}{\approx} \\
\mathcal{N}\Big(\ve(\rvx)\Big| \bar{\ve}, \textrm{Diag}\Big(\frac{1}{M}\sum_{m=1}^M \ve_m^2 - \bar{\ve}^2\Big) + \sigma^2 I\Big) 
=:\, q_{\textrm{semantic}}(\vz)
\end{align*}

where $\bar{\ve} = \frac{1}{M} \sum_{m=1}^M\ve_m, \; \ve_m = c_{\phi}(g_{\theta_m}(\vz)), \; \theta_m \sim q(\theta | \mathcal{D})$ and $c_{\phi}$ is a pre-trained feature extractor of choice (e.g., CLIP). With step $(1)$ we again denote the MC approximation (Eq. \ref{eq:bma-approx}) and step $(2)$ denotes moment-matching (with a diagonal covariance approximation). While the resulting posterior predictive based on the semantic likelihood can not be used to generate samples (since the likelihood is over CLIP features $\rve(\rvx) \in \mathcal{S}$ and not data $\rvx \in \mathcal{X}$), we can still compute its entropy which we use as our final estimate of generative uncertainty $u(\vz) \approx H(q_{\textrm{semantic}}(\vz))$. As described in Section~\ref{sec:sem-lik} and in Algorithm \ref{algo:diff-gen-unc}, we first generate a sample using a pretrained model $g_{\hat{\theta}}$ and then use the semantic posterior predictive $q_{\textrm{semantic}}$ solely for uncertainty estimation.

At this point it is important to acknowledge that by changing the likelihood to the semantic one we depart from the traditional Bayesian framework where the same likelihood is used both for finding the posterior $q(\theta | \mathcal{D})$ as well as in the approximation of the posterior predictive. However, we would like to emphasize that image generation using modern diffusion models poses specific challenges, which to the best of our knowledge, have not been addressed within the Bayesian framework yet. One such challenge is due to (extremely) high-dimensional sample spaces. For example, in the case of ImageNet 256x256 the dimensionality is $\sim 2 \cdot 10^5$. Our use of feature extractor $c_{\phi}$ via the semantic likelihood reduces the dimensionality (down to $512$), potentially making the MC approximation using few samples ($M$) `easier'.  Another challenge is that using a larger number of MC samples is computational prohibitive, since every additional sample corresponds to generating a new sample with a diffusion model which is costly. 

We hope that our promising experimental results based on the semantic likelihood (using a few MC samples only) will encourage the Bayesian community to further investigate the choice of the suitable likelihood in high-dimensional spaces (such as those of natural images) and fill-in the potentially missing theoretical gaps (e.g., due to changing the likelihood `post-hoc').

\newpage
\section{Additional Results}

\subsection{Qualitative Comparison with BayesDiff}
\label{app:exp-bayesdiff}

\begin{figure*}[htbp]
    \centering
    \begin{minipage}{0.47\textwidth}
        \centering
        \includegraphics[width=\textwidth]{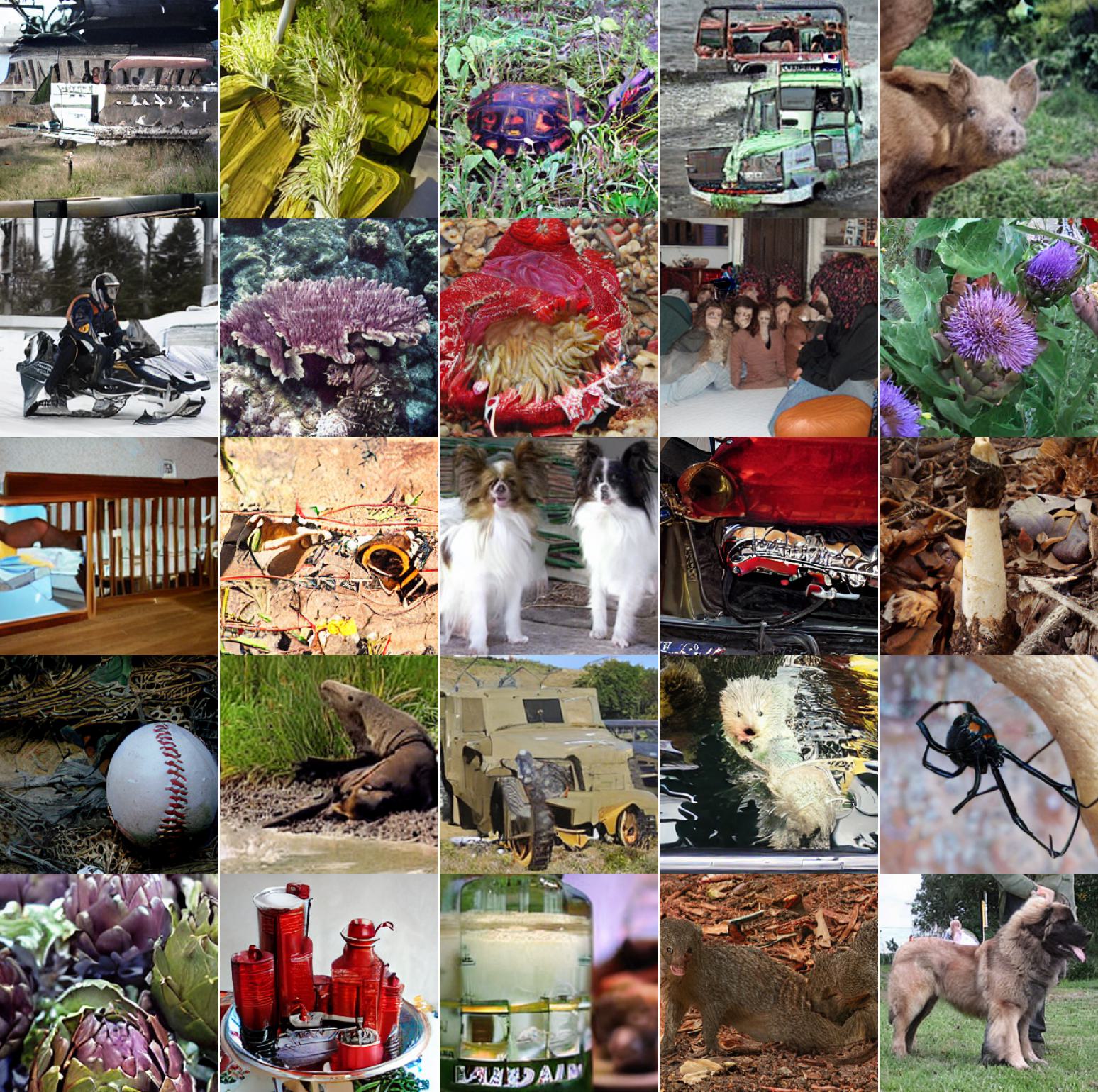}
    \end{minipage}
    \hfill
    \begin{minipage}{0.47\textwidth}
        \centering
        \includegraphics[width=\textwidth]{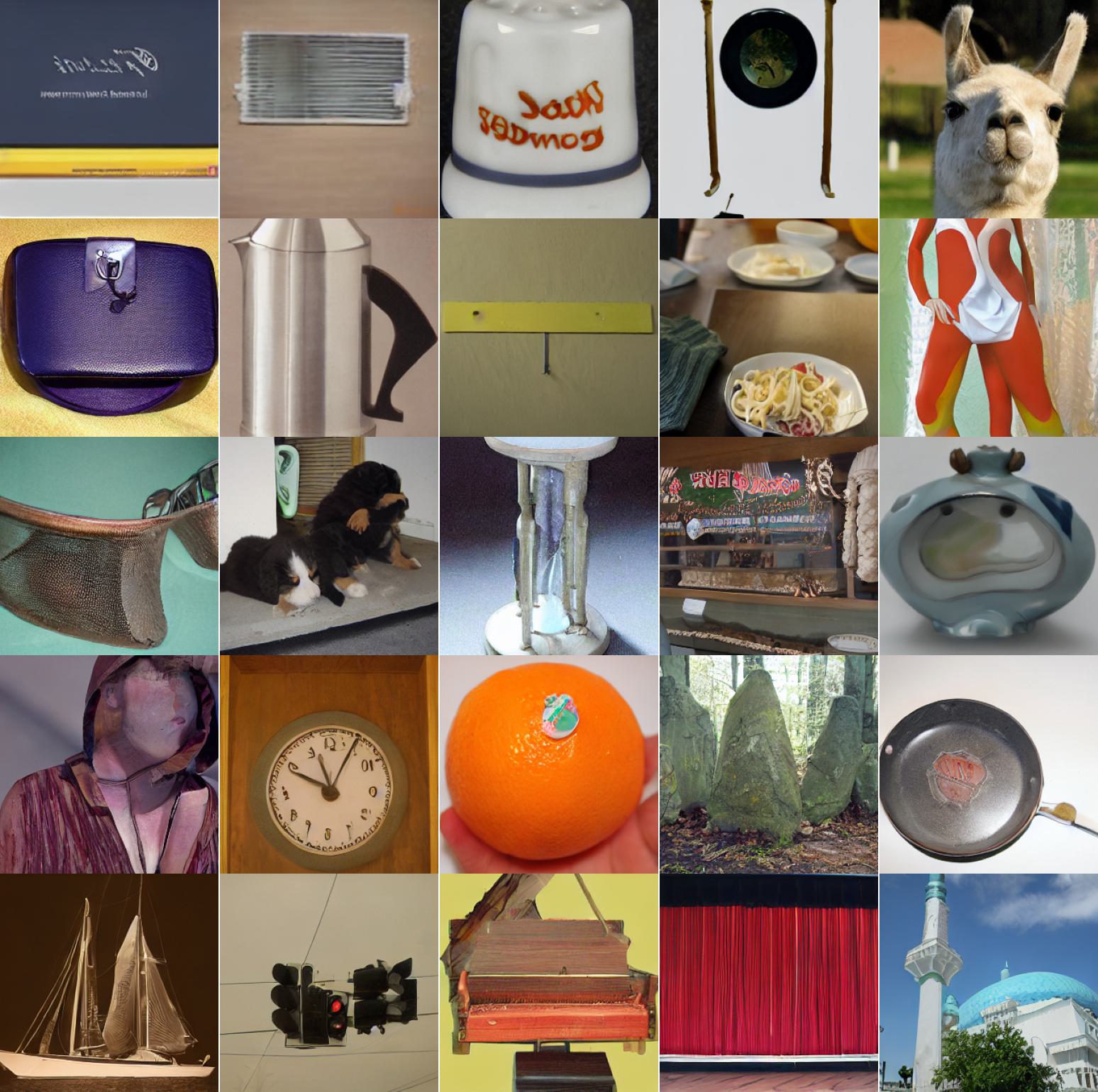}
    \end{minipage}
    \caption{Images with the highest (\textit{left}) and the lowest  (\textit{right}) BayesDiff uncertainty among 12K generations using a UViT diffusion model \citep{bao2023all}. BayesDiff uncertainty correlates poorly with visual quality and is overly sensitive to the background pixels. Same set of 12K generated images is used as in Figure~\ref{fig:best-worst-ours} to ensure a fair comparison.}
    \label{fig:best-worst-bayesdiff}
\end{figure*}

To further highlight the differences between our generative uncertainty and BayesDiff \citep{kou2023bayesdiff}, we present samples with the highest and lowest uncertainty according to BayesDiff in Figure~\ref{fig:best-worst-bayesdiff}. These samples are drawn from the same set of 12K ImageNet images generated using the UViT model \citep{bao2023all} as in Figure~\ref{fig:best-worst-ours}. Notably, BayesDiff's uncertainty score appears highly sensitive to background pixels—images with high uncertainty tend to have cluttered backgrounds, while those with low uncertainty typically feature clear backgrounds. Furthermore, as reflected in BayesDiff’s poor performance in terms of FID and precision (see Figures~\ref{fig:unc-filter-uvit}\&\ref{fig:unc-filter-adm}), some low-uncertainty examples exhibit noticeable artefacts, whereas certain high-uncertainty samples are of rather high-quality. For example, the image of a dog in the bottom-right corner of the high-uncertainty grid in Figure~\ref{fig:best-worst-bayesdiff} looks quite good despite being assigned (very) high uncertainty.

\begin{figure*}[htbp]
    \centering
    \begin{minipage}{0.47\textwidth}
        \centering
        \includegraphics[width=\textwidth]{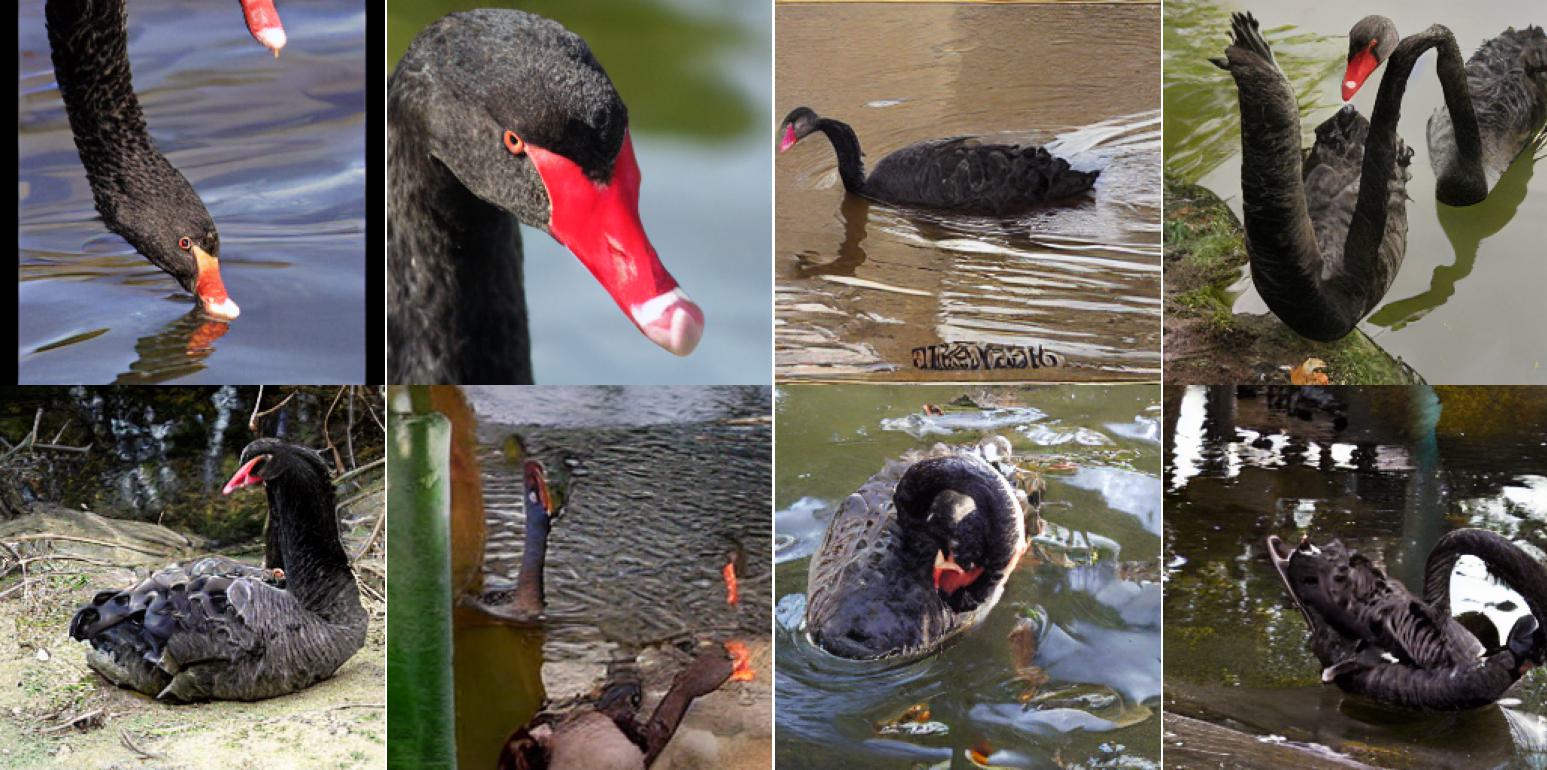}
    \end{minipage}
    \hfill
    \begin{minipage}{0.47\textwidth}
        \centering
        \includegraphics[width=\textwidth]{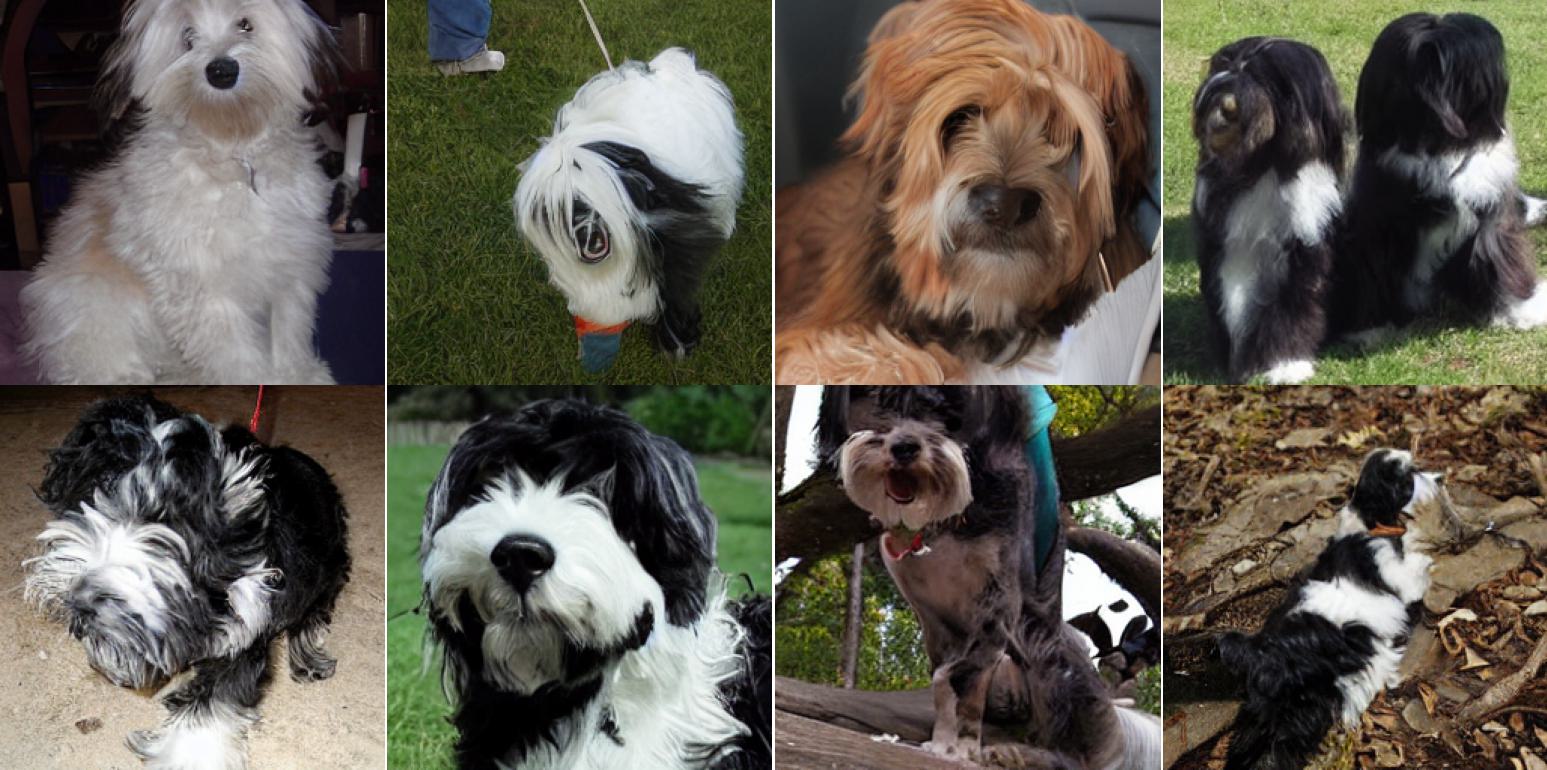}
    \end{minipage}
    \caption{Images with the highest (\textit{bottom}) and the lowest (\textit{top}) BayesDiff uncertainty among 128 generations using a UViT diffusion model for 2 classes: \texttt{black swan} (\textit{left}) and \texttt{Tibetan terrier} (\textit{right}). Same set of 128 generated images per class is used as in Figure~\ref{fig:best-worst-ours-cond} to ensure a fair comparison.}
    \label{fig:best-worst-bayesdiff-cond}
\end{figure*}

Similarly, in Figure~\ref{fig:best-worst-bayesdiff-cond}, we show low- and high-uncertainty samples according to BayesDiff for the same set of 128 images per class as in Figure~\ref{fig:best-worst-ours-cond}. Once again, we observe that BayesDiff’s uncertainty metric is less informative regarding a sample's visual quality compared to our generative uncertainty.

\newpage

\subsection{Ablation on Semantic Likelihood}
\label{app:exp-semlik}

\begin{figure*}[htbp]
    \centering
    \begin{minipage}{0.47\textwidth}
        \centering
        \includegraphics[width=\textwidth]{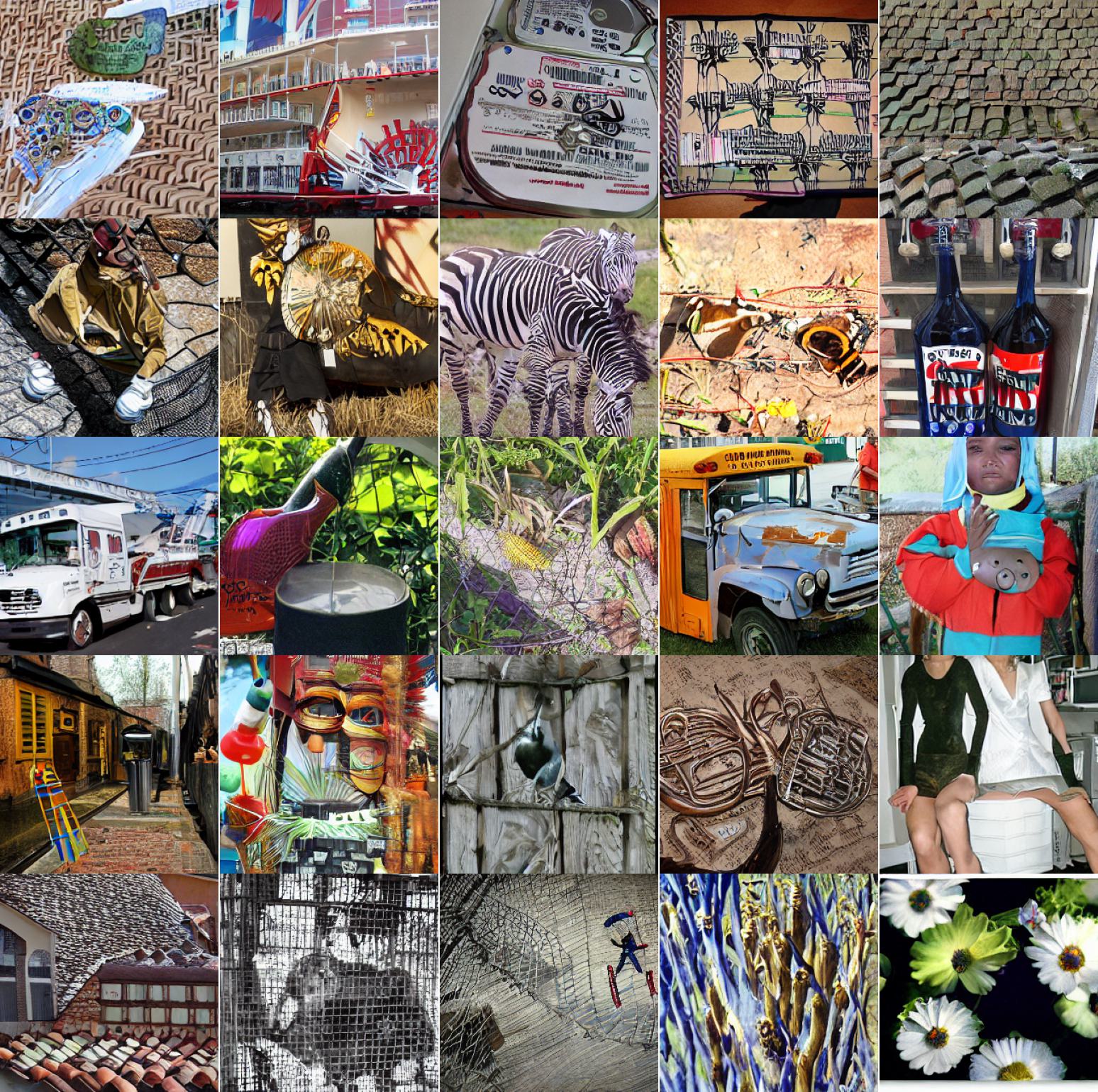}
    \end{minipage}
    \hfill
    \begin{minipage}{0.47\textwidth}
        \centering
        \includegraphics[width=\textwidth]{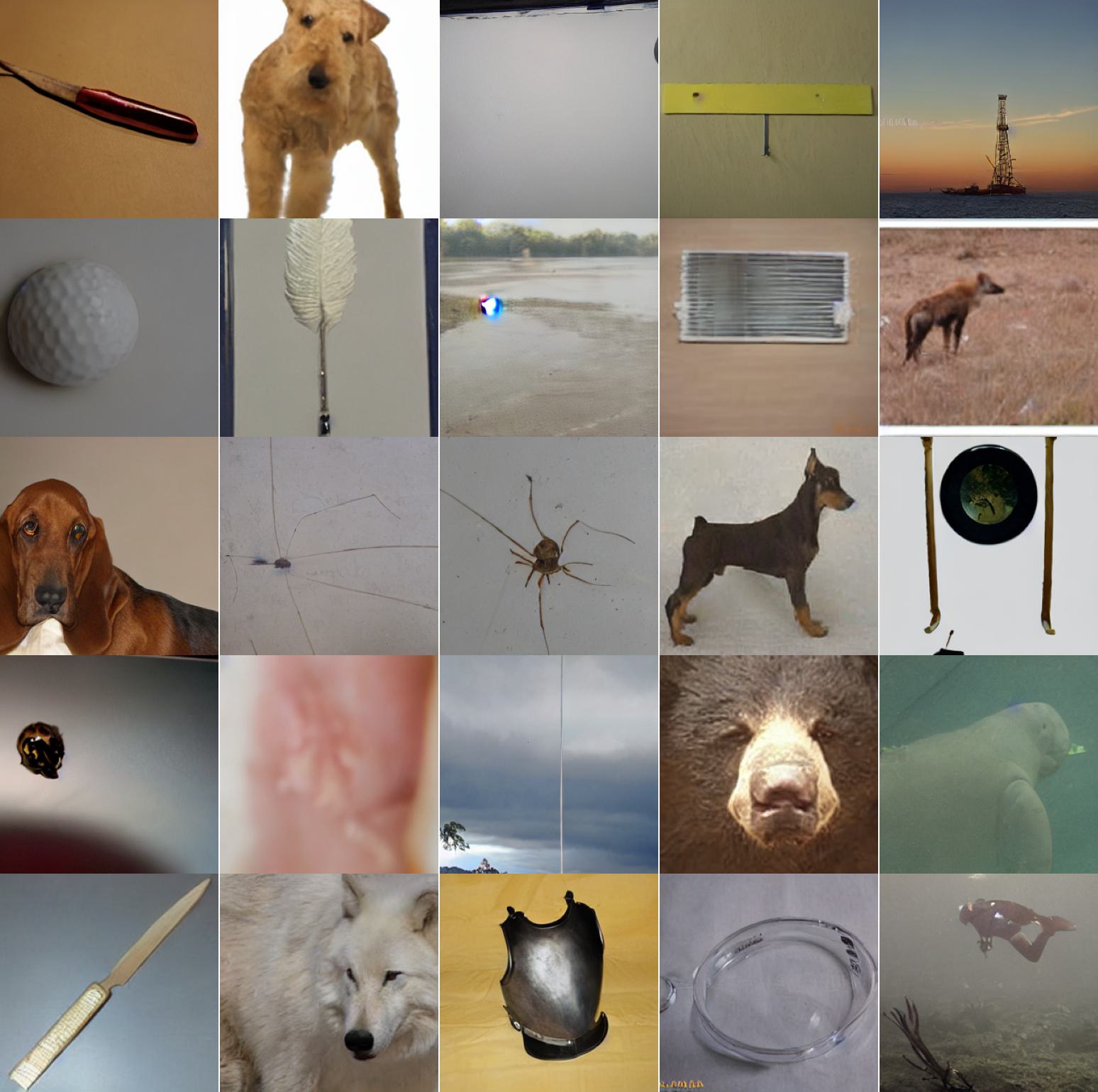}
    \end{minipage}
    \caption{Images with the highest (\textit{left}) and the lowest  (\textit{right}) `pixel-space' generative uncertainty among 12K generations using a UViT diffusion model. Pixel-space uncertainty correlates poorly with visual quality and is overly sensitive to the background pixels. Same set of 12K generated images is used as in Figure~\ref{fig:best-worst-ours} to ensure a fair comparison.}
    \label{fig:best-worst-ours-no-cphi}
\end{figure*}

To highlight the importance of using a semantic likelihood (Section~\ref{sec:sem-lik}) when leveraging uncertainty to detect low-quality generations, we conduct an ablation study in which we replace it with a standard Gaussian likelihood applied directly in pixel space (Eq.~\ref{eq:pixel-likelihood}). Figure~\ref{fig:best-worst-ours-no-cphi} presents the highest and lowest uncertainty images according to this `pixel-space' generative uncertainty. Notably, pixel-space uncertainty is overly sensitive to background pixels, mirroring the issue observed in BayesDiff (see Appendix~\ref{app:exp-bayesdiff}). This highlights the necessity of using semantic likelihood to obtain uncertainty estimates that are truly informative about the visual quality of generated samples. 

\subsection{Pixel-Wise Uncertainty}
\label{app:pixel}
\begin{figure}[ht]
    \centering
    \includegraphics[width=0.8\linewidth]{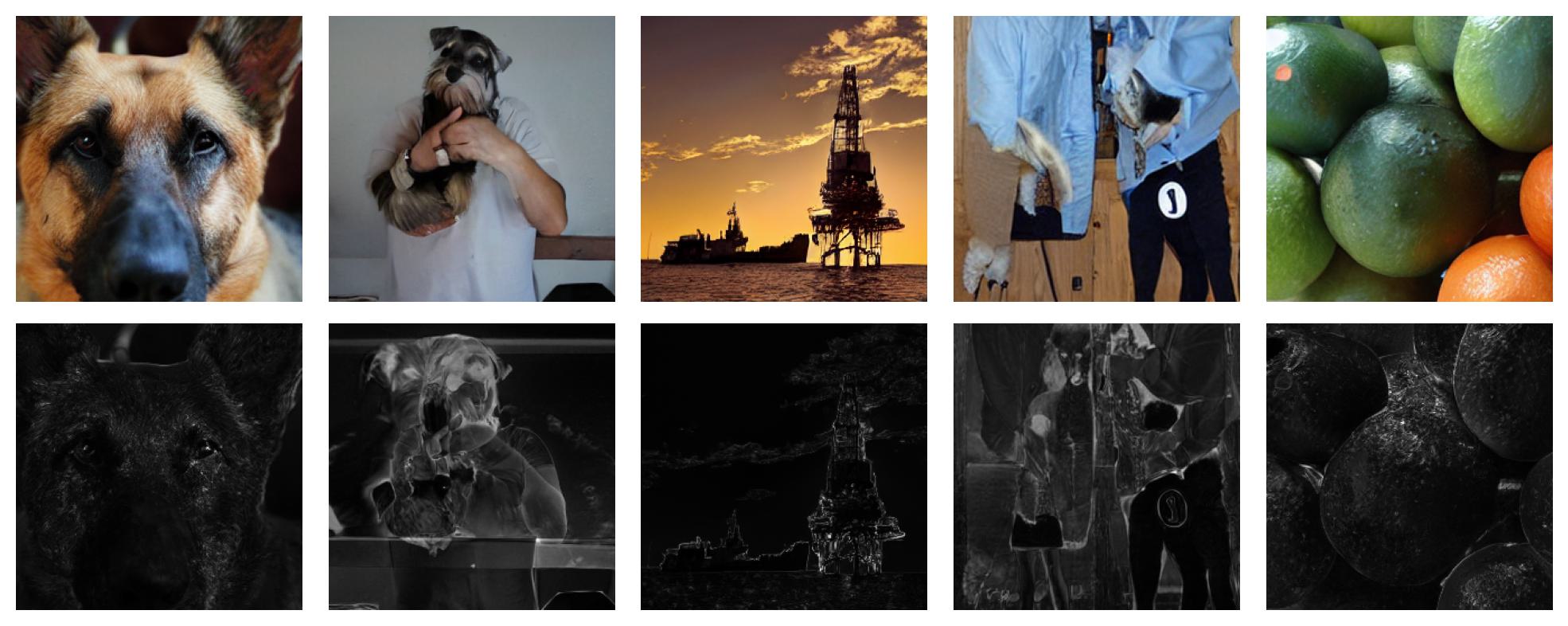}
    \caption{ Pixel-wise uncertainty based on our generative uncertainty for 5 generated samples using UViT diffusion.}
    \label{fig:pixel-unc}
\end{figure}

While not the primary focus of our work, we demonstrate how our generative uncertainty framework (Algorithm~\ref{algo:diff-gen-unc}) can be adapted to obtain pixel-wise uncertainty estimates. This is achieved by replacing our proposed semantic likelihood (Eq.~\ref{eq:sem-lik}) with a standard `pixel-space' likelihood (Eq.~\ref{eq:pixel-likelihood}). Figure~\ref{fig:pixel-unc} illustrates pixel-wise uncertainty estimates for 5 generated samples.

Although pixel-wise uncertainty received significant attention in past work \citep{kou2023bayesdiff, chanestimating, de2024diffusion}, there is currently no principled method for evaluating its quality. Most existing approaches rely on qualitative inspection, visualizing pixel-wise uncertainty for a few generated samples (as we do in Figure~\ref{fig:pixel-unc}). This further motivates our focus on sample-wise uncertainty estimates, where more rigorous evaluation frameworks—such as improvements in FID and precision on a set of filtered images—enable more meaningful comparisons between different approaches. 

\subsection{Comparison with Likelihood}
\label{app:llhood}
\begin{figure*}[htbp]
    \centering
    \begin{minipage}{0.47\textwidth}
        \centering
        \includegraphics[width=\textwidth]{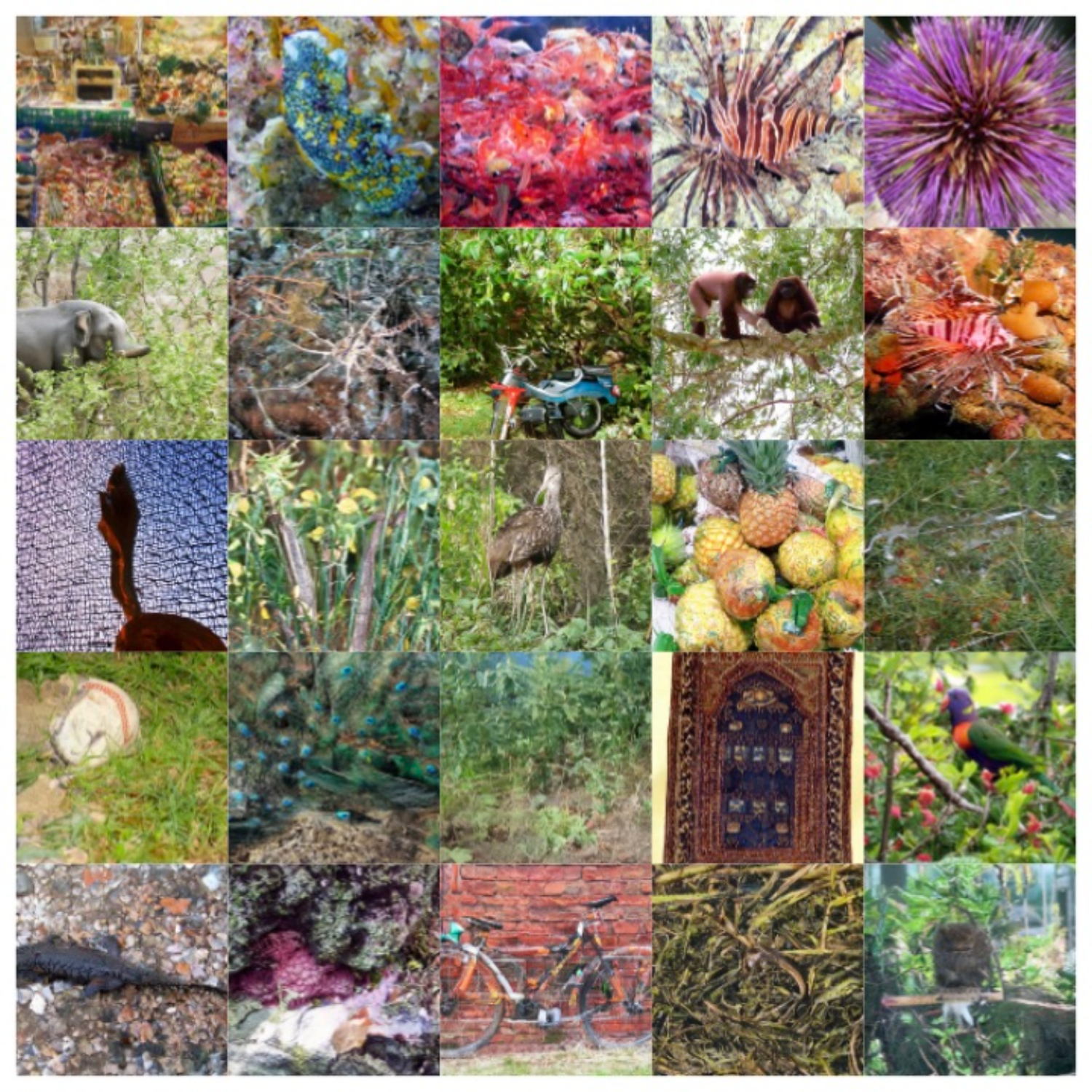}
    \end{minipage}
    \hfill
    \begin{minipage}{0.47\textwidth}
        \centering
        \includegraphics[width=\textwidth]{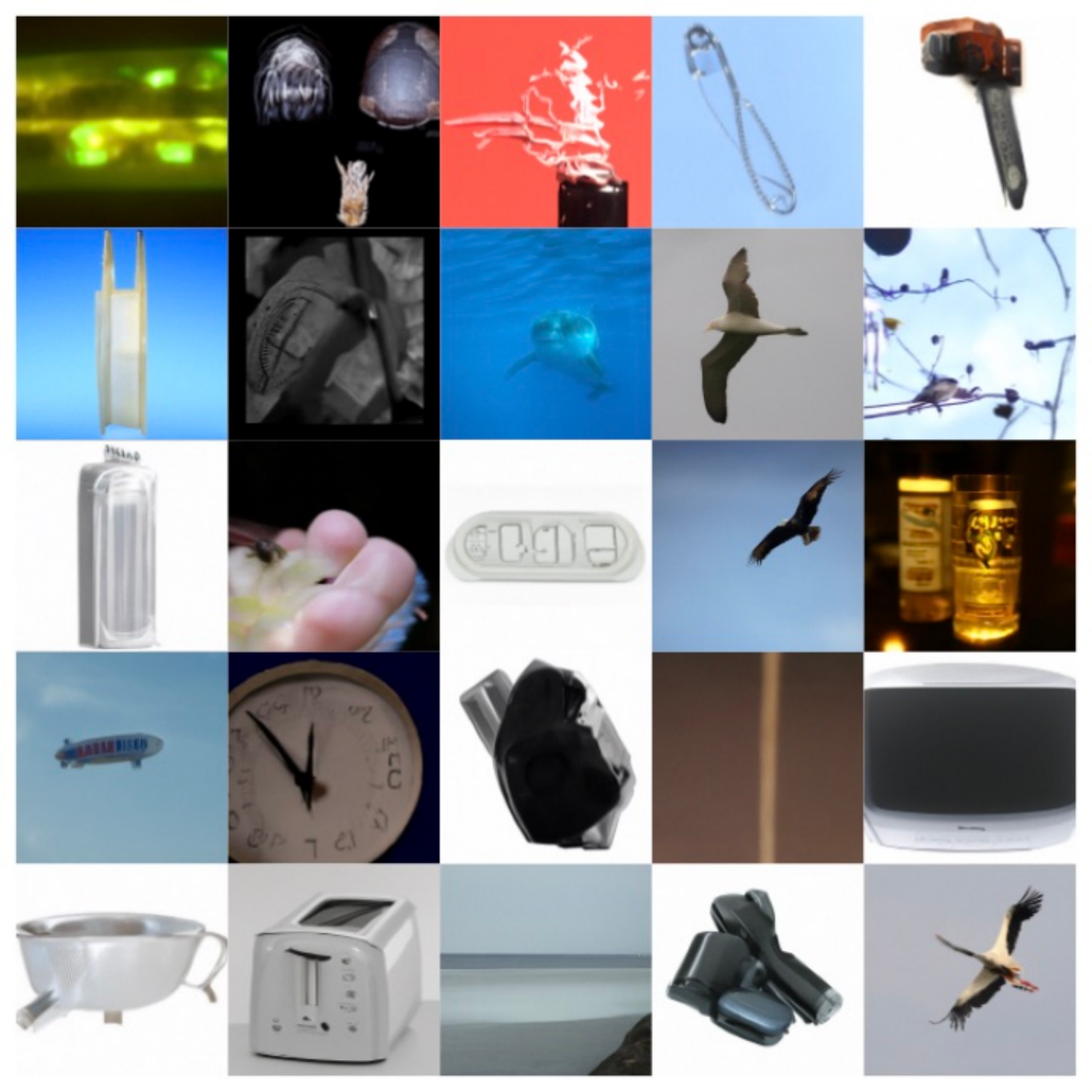}
    \end{minipage}
    \caption{The 25 `worst' (\textit{left}) and `best' (\textit{right}) samples generated by ADM trained on ImageNet 128x128 selected by lowest and highest likelihood among 12K generations.}
    \label{fig:best-worst-ours-likelihood}
\end{figure*}

We compare our generative uncertainty filtering criterion with a likelihood selection approach on the 12K images generated by ADM trained on ImageNet 128x128. Here retain the $n=10\textrm{K}$ generated images with highest likelihood. We utilize the implementation in  \cite{dhariwal2021diffusion} to compute the bits-per-dimension of each sample (one-to-one with likelihood). The 25 samples with lowest and highest likelihood are shown in Figure~\ref{fig:best-worst-ours-likelihood}. Visually, the likelihood objective heavily prefers simple images with clean backgrounds and not necessarily image quality. Note that this is consistent with other works that have reported likelihood to be an inconsistent identifier of image quality \citep{theis2016noteevaluationgenerativemodels, theis2024makes}. 
Quantitative results for image quality were consistent with our qualitative observations. The FID, precision, and recall for the best 10K images according to bits-per-dimension were 
$11.86 \pm 0.0026$, $58.23 \pm 0.02160$, and $70.45 \pm 0.0237$ over three runs. By point estimate, all three metrics are worse or indistinguishable from the Random baseline ($11.31\pm 0.07, 58.90 \pm 0.36, 70.68\pm 0.38$).


\newpage

\subsection{Comparison with Realism \& Rarity}
\label{app:corr}

\begin{figure*}[htbp]
\vspace{-5mm}
    \centering\includegraphics[width=0.99\textwidth]{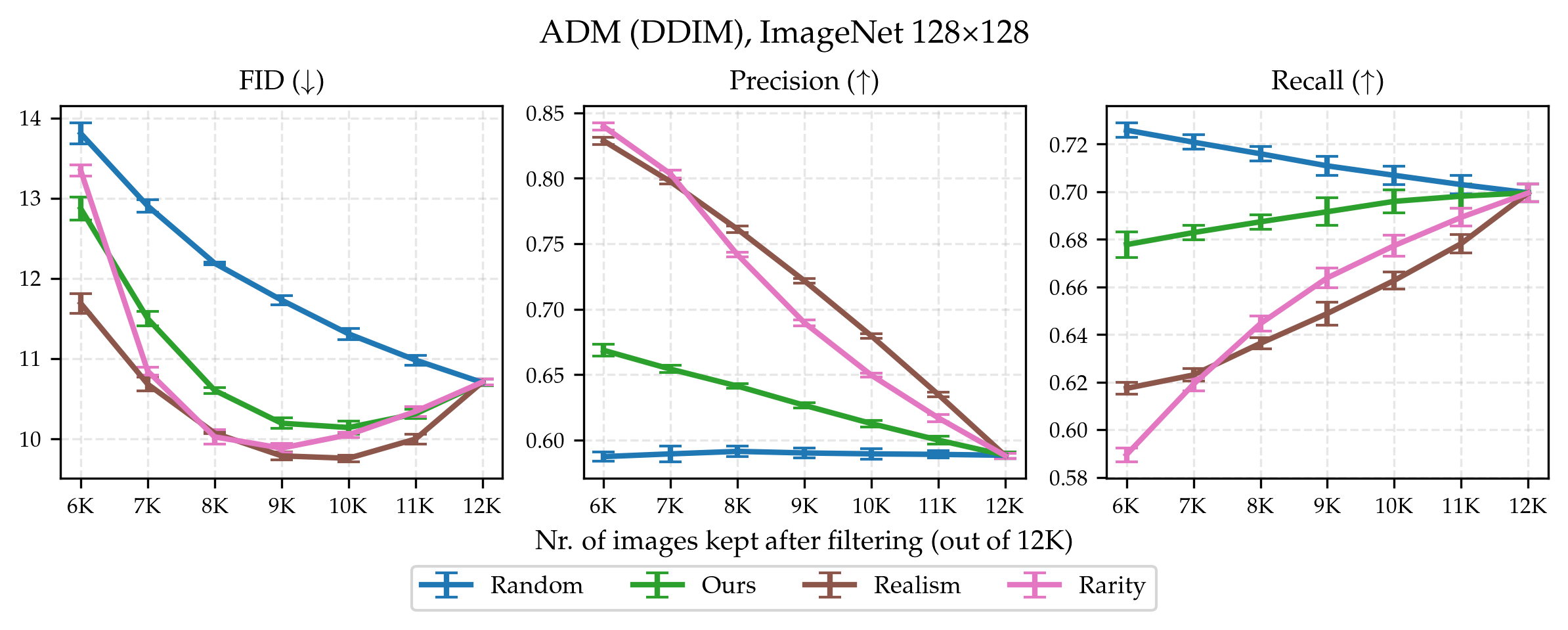}

    \caption{Image generation results for $n \in \{6\textrm{K}, 7\textrm{K}, \ldots, 11\textrm{K} \} $ filtered samples (out of 12K) for ADM diffusion model \citep{dhariwal2021diffusion}. Our generative uncertainty performs on par with realism \citep{kynkaanniemi2019improved} and rarity scores \citep{han2022rarity} in terms of FID (\emph{left}), while exhibiting a weaker precision-recall trade-off (\emph{middle} and \emph{right}). We report mean values along with standard deviations over 5 runs with different random seeds.}
    \label{fig:unc-filter-adm-realism-rarity}
\end{figure*}

\begin{figure*}[htbp]
\vspace{-5mm}
    \centering\includegraphics[width=0.99\textwidth]{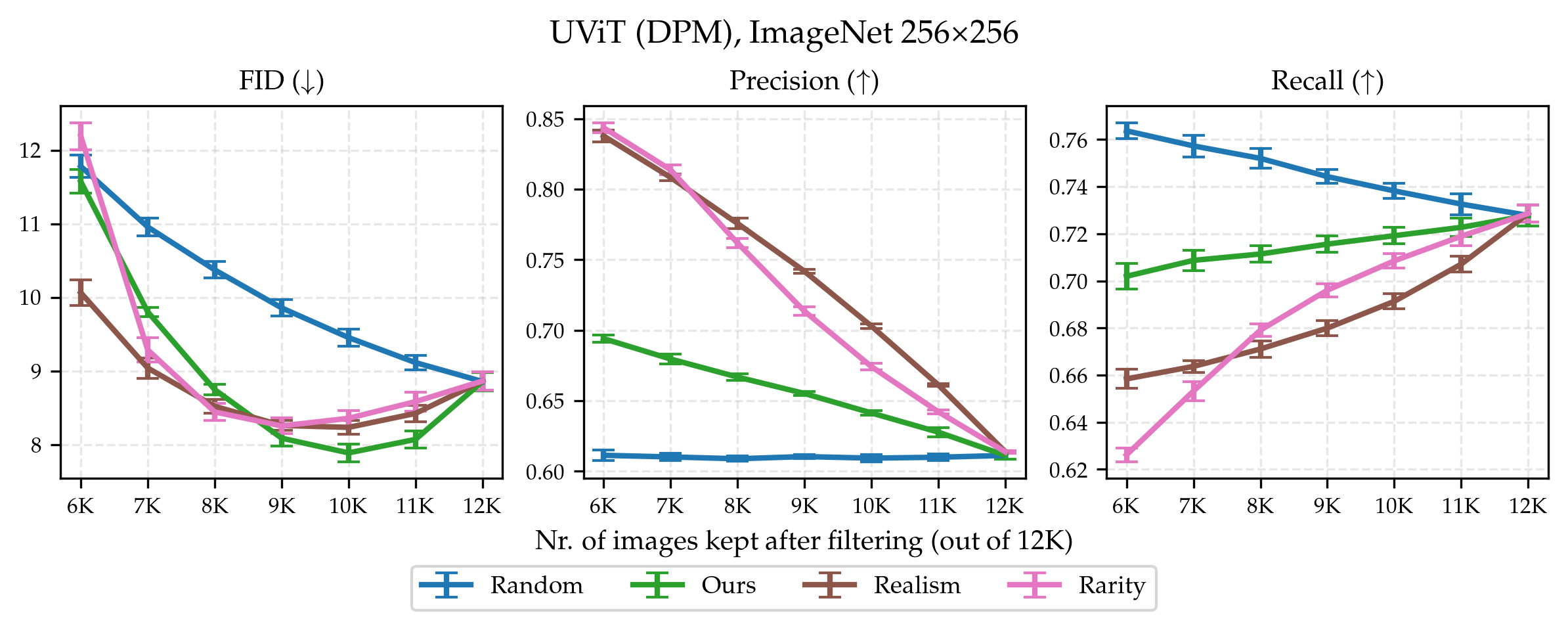}

    \caption{Image generation results for $n \in \{6\textrm{K}, 7\textrm{K}, \ldots, 11\textrm{K} \} $ filtered samples (out of 12K) for UViT diffusion model \citep{bao2023all}. Our generative uncertainty performs on par with realism \citep{kynkaanniemi2019improved} and rarity scores \citep{han2022rarity} in terms of FID (\emph{left}), while exhibiting a weaker precision-recall trade-off (\emph{middle} and \emph{right}). We report mean values along with standard deviations over 5 runs with different random seeds.}
    \label{fig:unc-filter-uvit-realism-rarity}
\end{figure*}

\begin{figure}[htbp]
    \centering
    \begin{minipage}{0.47\textwidth}
        \centering
        \includegraphics[width=0.8\textwidth]{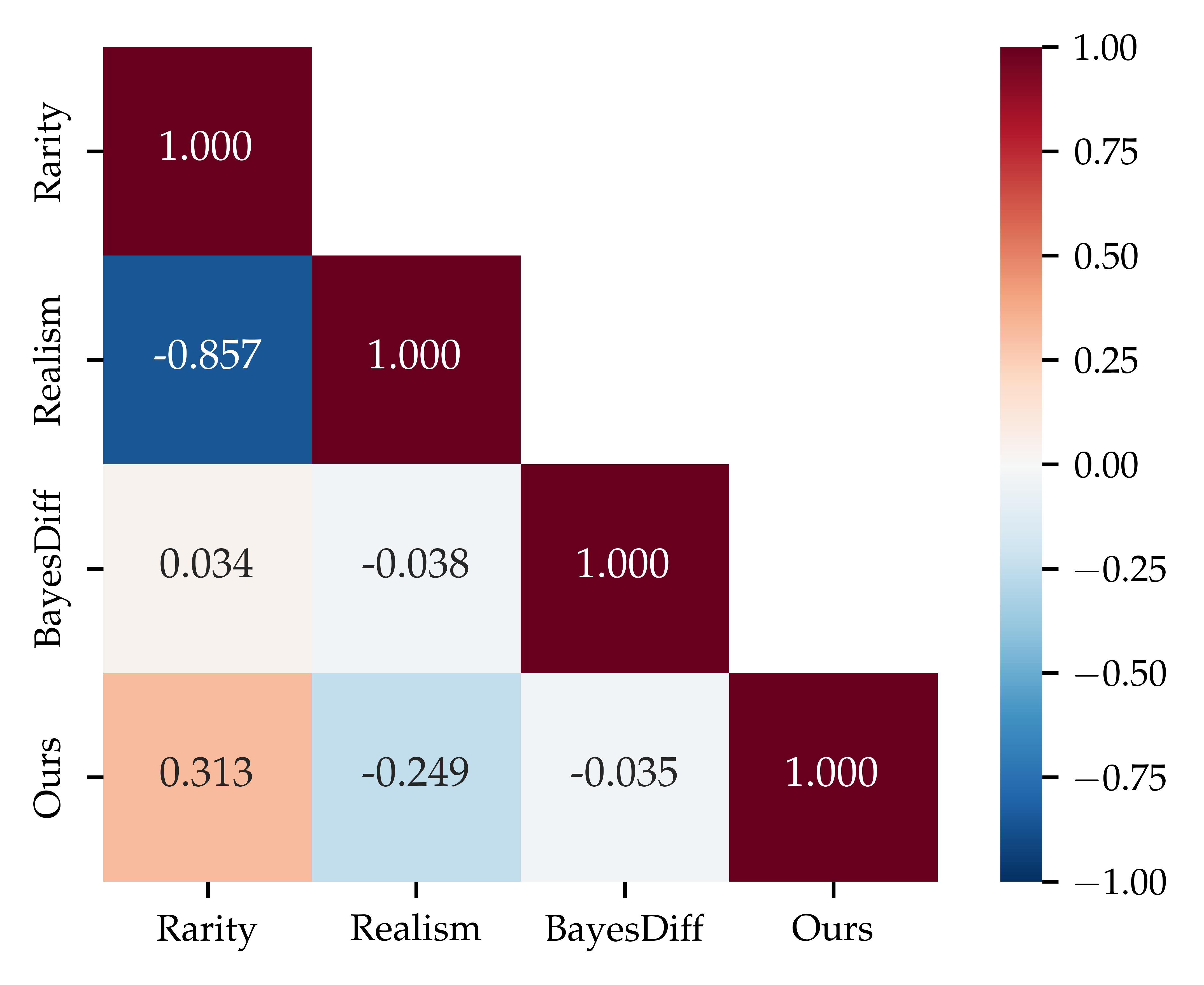}
    \end{minipage}
    \hfill
    \begin{minipage}{0.47\textwidth}
        \centering
        \includegraphics[width=0.8\textwidth]{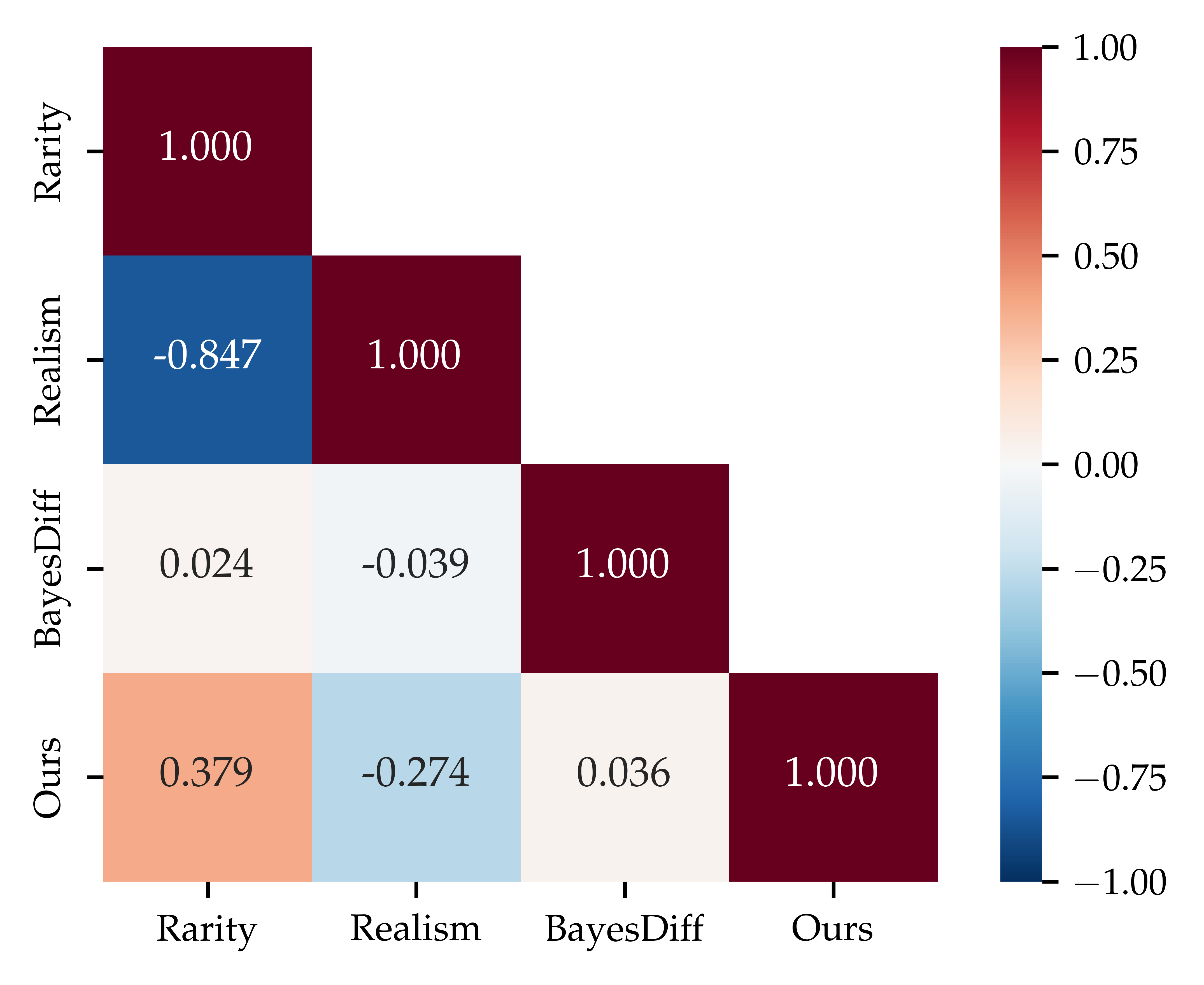}
    \end{minipage}
    \caption{Spearman correlation coefficient between different sample quality metrics for 12K ImageNet images generated using ADM (\textit{left}) and UViT (\textit{right}).}
    \label{fig:metrics-corr}
\end{figure}

To better understand the relationship between our generative uncertainty and non-uncertainty-based approaches such as realism \citep{kynkaanniemi2019improved} and rarity \citep{han2022rarity} scores, we compute the Spearman correlation coefficient between different sample-level metrics on a set of 12K generated images from the experiment in Section~\ref{sec:exp-poor}. As shown in Figure~\ref{fig:metrics-corr}, realism and rarity scores exhibit a strong correlation ($<-0.8$). This is unsurprising, as both scores are derived from the distance of a generated sample to a data manifold obtained using a reference dataset (e.g., a subset of training data or a separate validation dataset).\footnote{Such distance-based approaches are also commonly used to estimate prediction's quality in predictive models; see, for example, \cite{van2020uncertainty}.} 

In contrast, our generative uncertainty exhibits a weaker correlation ($<0.4$) with both realism and rarity scores. We attribute this to the fact that our uncertainty primarily reflects the limited training data used in training diffusion models (i.e., epistemic uncertainty), rather than the distance to a reference dataset, as is the case for realism and rarity scores.

Next, we investigate whether combining different scores can improve the detection of low-quality generations. When combining two scores, we first rank the 12K images based on each score individually, then compute the combined ranking by summing the two rankings and re-ranking accordingly. The results, shown in Table~\ref{tab:combined-scores}, indicate that combining realism and rarity leads to minor or no improvements in FID ($9.81$ compared to $9.76$ for realism alone on ADM. However, combining our generative uncertainty with either realism or rarity achieves the best FID performance ($9.54$ on ADM). These results suggest that ensembling scores that capture different aspects of generated sample quality is a promising direction for future research.

\begin{table*}[htbp]
    \centering
    \caption{Image generation results for $n=10\textrm{K}$ filtered samples (out of $12\textrm{K}$) based on combined metrics. Combining our generative uncertainty outperforms combining realism and recall in terms of FID. We report mean values along with standard deviation over 5 runs with different random seeds.}
    \label{tab:combined-scores}
    \begin{tabular}{lccc|ccc}
        \toprule
        & \multicolumn{3}{c}{ADM (DDIM), ImageNet 128×128} & \multicolumn{3}{c}{UViT (DPM), ImageNet 256×256} \\
        \cmidrule(lr){2-4} \cmidrule(lr){5-7}
         & \textbf{FID} $(\downarrow)$  & \textbf{Precision} $(\uparrow)$ & \textbf{Recall} $(\uparrow)$ & \textbf{FID} $(\downarrow)$ & \textbf{Precision} $(\uparrow)$ & \textbf{Recall} $(\uparrow)$ \\
        \midrule
        \textbf{Realism + Rarity} & $9.81 \pm 0.06$ & $67.06 \pm 0.29$ & $66.73 \pm 0.37$  & $8.26 \pm 0.07$  & $69.01 \pm 0.33$ & $69.86 \pm 0.36$  \\
        \textbf{Ours+ Realism} & $9.54 \pm 0.04$  & $66.41 \pm 0.15$ & $67.04 \pm 0.47$ & $7.60 \pm 0.10$ &  $68.33 \pm 0.09$ & $69.75 \pm 0.42$\\
        \textbf{Ours + Rarity}  & $9.56 \pm 0.06$ & $65.44 \pm 0.26$ & $67.36 \pm 0.54$ & $7.56 \pm 0.12$ & $67.48 \pm 0.18$ & $70.18 \pm 0.40$ \\
        \bottomrule
    \end{tabular}
\end{table*}

\newpage

\subsection{Class-averaged generative uncertainty}
\label{app:class-ent}

\begin{wrapfigure}{r}{0.5\textwidth} 
    \centering
    \vspace{-1\baselineskip}
    \includegraphics[width=0.48\textwidth]{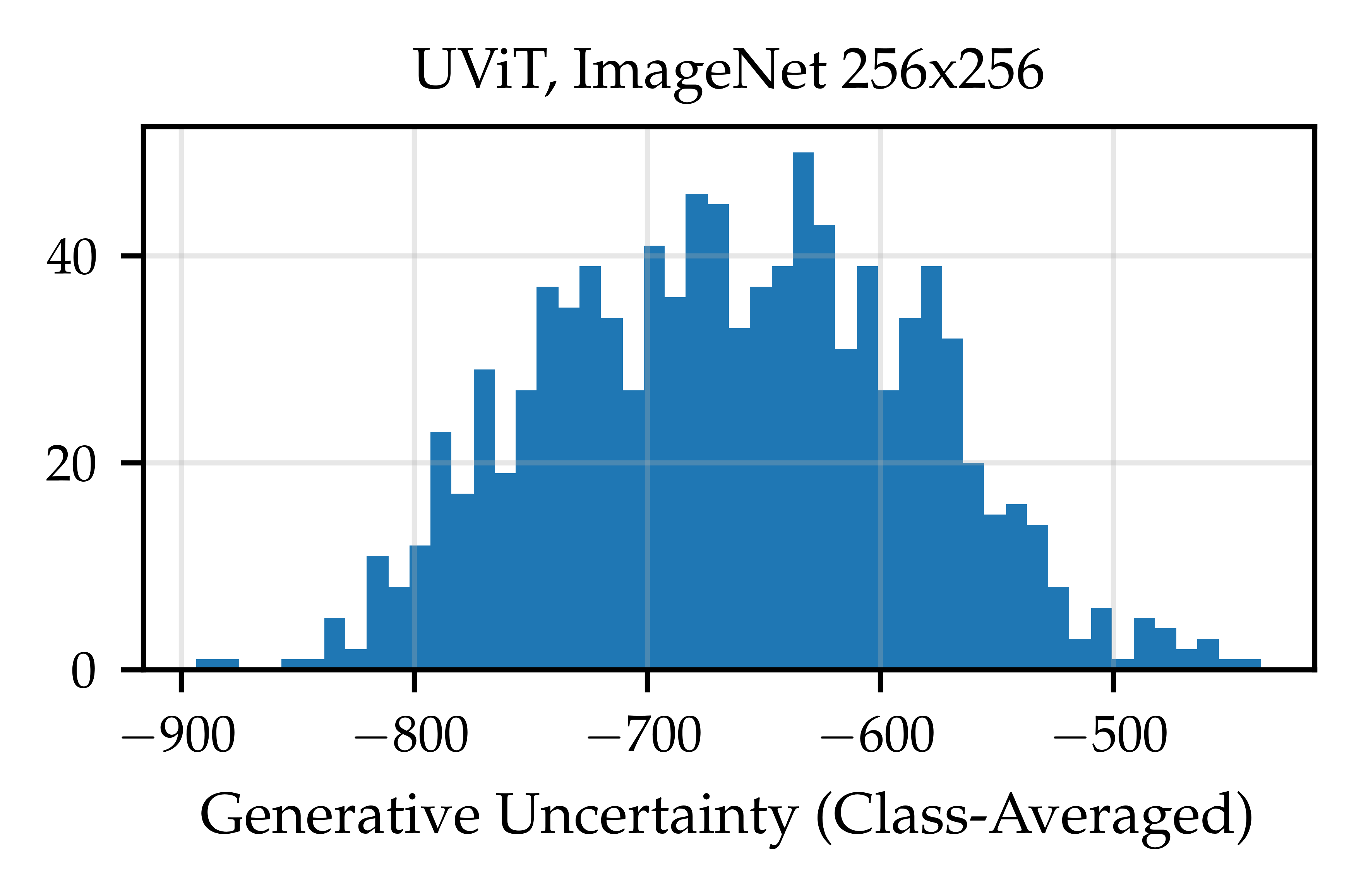}
    \caption{A histogram of class-averaged generative uncertainties for 12K generated samples using UViT.}
    \label{fig:avg-ent}
\vspace{-1\baselineskip}
\end{wrapfigure}

To better understand the drop in sample diversity (recall) when using our generative uncertainty to filter low-quality samples in Figures~\ref{fig:unc-filter-uvit}\&\ref{fig:unc-filter-adm}, we analyze the distribution of average entropy per conditioning class. Specifically, for each of the 12K generated images, we randomly sample a conditioning class to mimic unconditional generation. As a result, all 1,000 ImageNet classes are represented among the 12K generated samples. Next, we compute our generative uncertainty for each sample and then average the uncertainties within each class. A plot of class-averaged uncertainties is shown in Figure~\ref{fig:avg-ent}. Since class-averaged uncertainties exhibit considerable variance, the class distribution in the 10K filtered samples deviates somewhat from that of the original 12K images, thereby explaining the reduction in diversity (recall).

While our primary focus in this work is on providing per-sample uncertainty estimates $u(\vz)$, we can also obtain uncertainty estimates for the conditioning variable $u(\vy)$ (e.g., a class label), by averaging over all samples corresponding to a particular $\vy \in \Y$ as done in Figure~\ref{fig:avg-ent}. These estimates resemble the epistemic uncertainty scores proposed in DECU \citep{berryshedding} and could be used to identify conditioning variables for which generated samples are likely to be of poor quality. We leave further exploration of generative uncertainty at the level of conditioning variables for future work.

\vspace{-4mm}
\subsection{Flow Matching}
\label{app:flow-match}
\vspace{-2mm}

\begin{figure*}[htbp]
    \centering
    \begin{minipage}{0.45\textwidth}
        \centering
        \includegraphics[width=\textwidth]{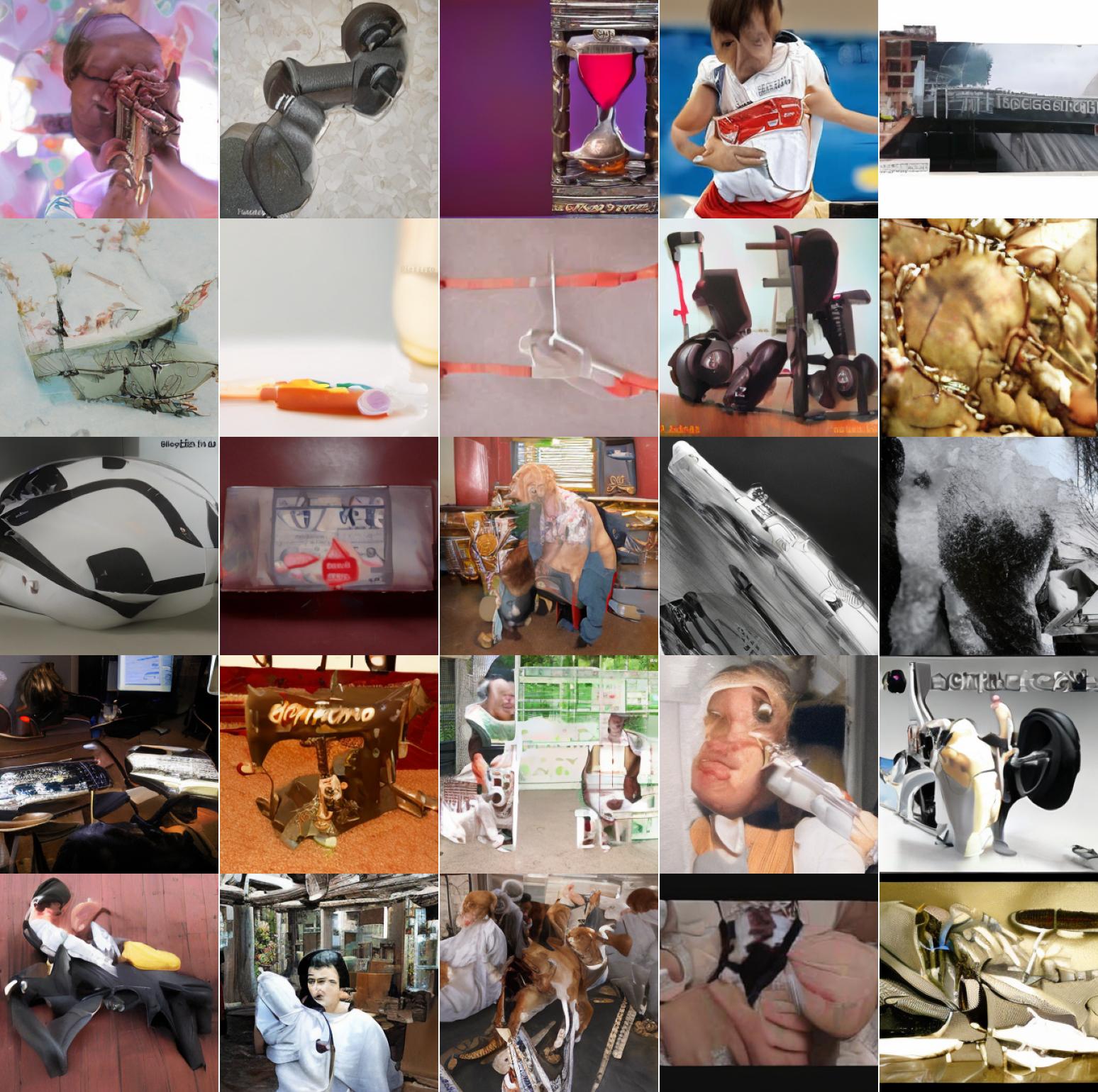}
    \end{minipage}
    \hfill
    \begin{minipage}{0.45\textwidth}
        \centering
        \includegraphics[width=\textwidth]{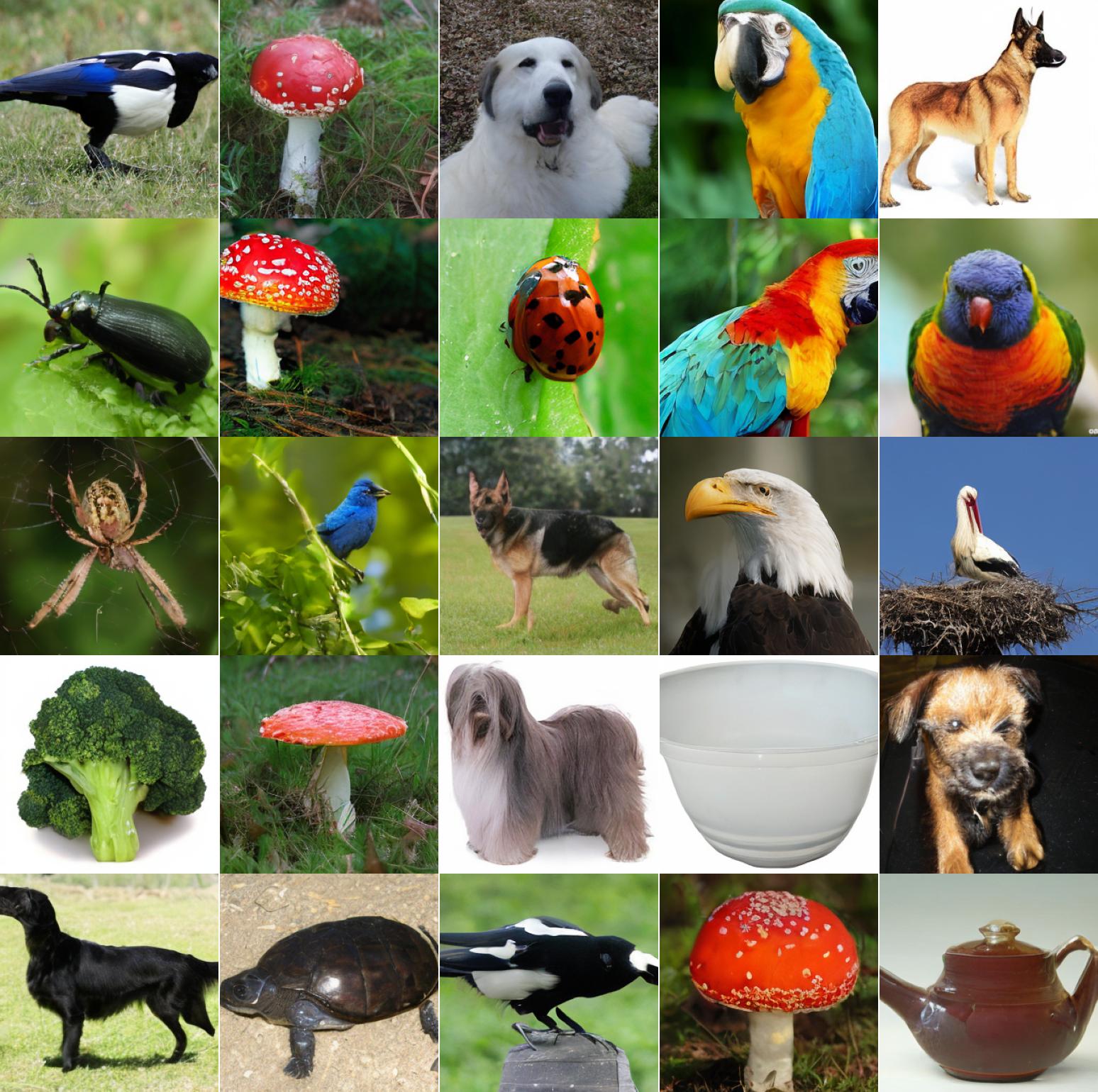}
    \end{minipage}
    \caption{Images with the highest (\textit{left}) and the lowest  (\textit{right}) generative uncertainty among 12K generations using a latent flow matching model \citep{dao2023flow}. Uncertainty correlates with visual quality, as high-uncertainty samples exhibit numerous artefacts, whereas low-uncertainty samples resemble canonical images of their respective conditioning class.}
    \label{fig:best-worst-ours-lfm}
\end{figure*}

To demonstrate that our generative uncertainty framework (Section~\ref{sec:methods}) extends beyond diffusion models, we apply it here to the recently popularized flow matching approach \citep{lipman2022flow, liu2022flow, albergo2023stochastic}. Specifically, we consider a latent flow matching formulation \citep{dao2023flow} with a DiT backbone \citep{peebles2023scalable}. For sampling, we employ a fifth-order Runge-Kutta ODE solver (\texttt{dopri5}). In Figure~\ref{fig:best-worst-ours-lfm}, we illustrate the samples with the highest and lowest generative uncertainty among 12K generated samples. On a filtered set of 10K images, our generative uncertainty framework achieves an FID of $10.48$ and a precision of $64.71$, significantly outperforming a random baseline, which yields an FID of $11.80$ and a precision of $61.04$.

\newpage

\section{Implementation Details}
\label{app:impl}

All our experiments can be conducted on a single $A100$ GPU, including the fitting of the Laplace posterior (Section~\ref{sec:methods-llla}). Our code is publicly available at \url{https://github.com/metodj/DIFF-UQ}.

\begin{wraptable}{r}{0.5\textwidth}
    \centering
    \setlength{\tabcolsep}{6pt}  
    \begin{tabular}{lccc}
        \toprule
        \textbf{} & \textbf{All Params.} & \textbf{LL Params.} & \textbf{LL Name} \\
        \midrule
        \textbf{ADM} & $\sim 421 \times 10^6$ & $\sim 14 \times 10^3$ & \texttt{out.2} \\
        \textbf{UViT} & $\sim 500 \times 10^6$ & $\sim 18 \times 10^3$ & \texttt{decoder\_pred}\\
        \textbf{DiT} & $\sim 131 \times 10^6 $ & $\sim 1.2 \times 10^6 $ & \texttt{final\_layer} \\
        \bottomrule
    \end{tabular}
    \caption{Details of our last-layer (LL) Laplace approximation. The first column presents the total number of model parameters, while the second and third columns indicate the number of parameters in the last layer and its name, respectively}
    \label{tab:llla}
\end{wraptable}

\paragraph{Laplace Approximation} When fitting a last-layer Laplace approximation (Section~\ref{sec:methods-llla}), we closely follow the implementation from BayesDiff \citep{kou2023bayesdiff}. Specifically, we use the empirical Fisher approximation with a diagonal factorization for Hessian computation. As the prior, we adopt a simple isotropic Gaussian distribution, $p(\theta) = \mathcal{N}(0, \gamma^{-1} I)$. The prior precision parameter and observation noise are fixed at $\gamma = 1$ and $\sigma = 1$, respectively. We report ablations for both parameters in Tables~\ref{tab:abl-prior} and~\ref{tab:abl-sigma}, finding that neither has a significant impact on the results. For Hessian computation, we utilize $1\%$ of the training data for ImageNet 128×128 and $2\%$ for ImageNet 256×256. Further details about the last layer of each diffusion model are provided in Table~\ref{tab:llla}, where we observe that fewer than $1\%$ of the parameters receive a `Bayesian treatment'. We utilize \texttt{laplace}\footnote{\url{https://github.com/aleximmer/Laplace}} library  in our implementation.

As discussed in Section~\ref{sec:limit}, improving the quality of the Laplace approximation—such as incorporating both first and last layers instead of only the last layer \citep{daxberger2021bayesian, sharma2023bayesian} or optimizing Laplace hyperparameters (e.g., prior precision and observation noise) \citep{immer2021scalable}—could further enhance the quality of generative uncertainty and represents a promising direction for future work.

\paragraph{Sampling with Generative Uncertainty} For our main experiment in Section~\ref{sec:exp-poor}, we generate 12K images using the pretrained ADM model \citep{dhariwal2021diffusion} for ImageNet 128×128 and the UViT model \citep{bao2023all} for ImageNet 256×256. Following BayesDiff \citep{kou2023bayesdiff}, we use a DDIM sampler \citep{song2020denoising} for the ADM model and a DPM-2 sampler \citep{lu2022dpm} for the UViT model, both with $T=50$ sampling steps.

To compute generative uncertainty (Algorithm~\ref{algo:diff-gen-unc}), we first sample $M=5$ sets of weights from the posterior $q(\theta | \D)$. Then, for each of the initial 12K random seeds, we generate $M$ additional samples. The same set of model weights $\{\theta_m\}_{m=1}^M$ is used for all 12K samples for efficiency reasons. For semantic likelihood (Eq.~\ref{eq:sem-lik}), we use a pretrained CLIP encoder \citep{radford2021learning} and set the semantic noise to $\sigma^2=0.001$ .

\paragraph{Baselines} For all baselines, we use the original implementation provided by the respective papers, except for \citep{de2024diffusion}, which we reimplemented ourselves since we were unable to get their code to run. Moreover, we use the default settings (e.g., hyperparameters) recommended by the authors for all baselines. For realism \citep{kynkaanniemi2019improved} and rarity \citep{han2022rarity} we use InceptionNet \citep{szegedy2016rethinking} as a feature extractor and a subset of 50K ImageNet training images as the reference dataset. For samples where the rarity score is undefined (i.e., those that lie outside the estimated data manifold), we set it to \texttt{inf}.

\begin{table}[ht]
\centering
\begin{minipage}{0.48\linewidth}
\centering
\caption{Ablation on prior precision parameter $\gamma$. Results for ADM model on ImageNet 128x128 dataset based on $n=10\textrm{K}$ filtered images (out of 12K).}
\label{tab:abl-prior}
\begin{tabular}{@{}lccc@{}}
\toprule
$\gamma$ & \textbf{FID} $(\downarrow)$  & \textbf{Precision} $(\uparrow)$ & \textbf{Recall} $(\uparrow)$ \\
\midrule
$1.0$    & $10.04 \pm 0.14$ & $61.28 \pm 0.23$ & $69.55 \pm 0.49$ \\
$0.01$   & $10.04 \pm 0.12$ & $61.14 \pm 0.25$ & $69.59 \pm 0.52$ \\
$0.1$    & $10.05 \pm 0.09$ & $61.19 \pm 0.21$ & $69.75 \pm 0.45$ \\
$10.$    & $10.01 \pm 0.15$ & $60.95 \pm 0.28$ & $69.71 \pm 0.54$ \\
$100.$   & $10.06 \pm 0.11$ & $61.12 \pm 0.26$ & $69.62 \pm 0.50$ \\
\bottomrule
\end{tabular}
\end{minipage}
\hfill
\begin{minipage}{0.48\linewidth}
\centering
\caption{Ablation on likelihood noise $\sigma^2$ parameter. Results for ADM model on ImageNet 128x128 dataset based on $n=10\textrm{K}$ filtered images (out of 12K).}
\label{tab:abl-sigma}
\begin{tabular}{@{}lccc@{}}
\toprule
$\sigma^2$ & \textbf{FID} $(\downarrow)$  & \textbf{Precision} $(\uparrow)$ & \textbf{Recall} $(\uparrow)$ \\
\midrule
$0.1$      & $10.30 \pm 0.11$ & $60.62 \pm 0.31$ & $69.99 \pm 0.43$ \\
$0.01$     & $10.18 \pm 0.04$ & $61.18 \pm 0.38$ & $69.61 \pm 0.59$ \\
$0.001$    & $10.04 \pm 0.14$ & $61.28 \pm 0.23$ & $69.55 \pm 0.49$ \\
$0.0001$   & $10.01 \pm 0.14$ & $61.34 \pm 0.23$ & $69.50 \pm 0.39$ \\
$0.00001$  & $10.10 \pm 0.06$ & $61.40 \pm 0.28$ & $69.53 \pm 0.57$ \\
\bottomrule
\end{tabular}
\end{minipage}
\end{table}

\section*{Additional Acknowledgements and Disclaimers}
Parts of this research were supported by the Intelligence Advanced Research Projects Activity (IARPA) via Department of Interior/ Interior Business Center (DOI/IBC) contract number 140D0423C0075. The U.S. Government is authorized to reproduce and distribute reprints for Governmental purposes
notwithstanding any copyright annotation thereon. Disclaimer: The views and conclusions
contained herein are those of the authors and should not be interpreted as necessarily
representing the official policies or endorsements, either expressed or implied, of IARPA, DOI/IBC, or the U.S. Government.

\end{document}